%% file: journal.tex
\documentclass[10pt,letterpaper]{article}

\usepackage[margin=1.5in]{geometry}
\usepackage{graphicx}
\usepackage{amsfonts, amsmath, amssymb, bm, bbm, mathtools}
\usepackage[usenames,dvipsnames,svgnames,table]{xcolor}
\usepackage{url}
\usepackage{booktabs}
\usepackage{subfig, multirow}
\usepackage{chngpage}
\usepackage{natbib}
\usepackage[colorlinks=true, allcolors=black]{hyperref}             
\usepackage{mdwlist}
\usepackage[T1]{fontenc}
\usepackage{microtype}
\microtypecontext{spacing=nonfrench}
\usepackage{authblk}
\usepackage{newtxtext}
\usepackage{tablefootnote}
\usepackage[shortlabels]{enumitem}

\input{macros}

\providecommand{\keywords}[1]
{
  \small\noindent
  \textbf{Keywords:} #1
}

\date{}

\begin{document}
\title{\bfseries\Large Embracing the Disharmony in Medical Imaging: \\A Simple and Effective Framework for Domain Adaptation}
%
%
\author[1,3]{\normalsize Rongguang Wang\footnote{Corresponding author: \href{mailto:rgw@seas.upenn.edu}{rgw@seas.upenn.edu}, 3700 Hamilton Walk, 7th Floor, Center for Biomedical Image Computing and Analytics, University of Pennsylvania, Philadelphia, PA 19104, USA.}}
\author[1,2]{Pratik Chaudhari
}
\author[1,3,4]{Christos Davatzikos
}
\author[ ]{$^\dagger$}
\affil[1]{\small Department of Electrical and Systems Engineering, University of Pennsylvania, PA, USA}
\affil[2]{General Robotics, Automation, Sensing and Perception (GRASP) Laboratory, University of Pennsylvania}
\affil[3]{Center for Biomedical Image Computing and Analytics (CBICA), University of Pennsylvania}
\affil[4]{Department of Radiology, Perelman School of Medicine, University of Pennsylvania}
\affil[ ]{$^\dagger$for the iSTAGING consortium$^5$, the PHENOM consortium$^6$, and the Alzheimer’s Neuroimaging consortium$^7$ \footnote{See investigators in~\citet{habes2021brain} for $^5$, \citet{chand2020two} for $^6$, and~\citet{jack2008alzheimer} for $^7$.}}
\affil[ ]{Email: 
\href{mailto:rgw@seas.upenn.edu}{rgw@seas.upenn.edu},
\href{mailto:pratikac@seas.upenn.edu}{pratikac@seas.upenn.edu},
\href{mailto:christos.davatzikos@pennmedicine.upenn.edu}{christos.davatzikos@pennmedicine.upenn.edu}}
\maketitle              
%
\input{./journal_docs/abstract.tex}

\input{./journal_docs/introduction.tex}
\input{./journal_docs/related_work.tex}

\input{./journal_docs/method.tex}

\input{./journal_docs/experiments.tex}

\input{./journal_docs/results.tex}

\input{./journal_docs/conclusion.tex}

\section*{Acknowledgments}

This work was supported by the National Institute on Aging (grant numbers RF1AG054409 and U01AG068057) and the National Institute of Mental Health (grant number R01MH112070).
Pratik Chaudhari would like to acknowledge the support of cloud computing credits through the Amazon Machine Learning Research Award.

%
%
%

\newpage

\bibliographystyle{apalike}
\bibliography{refs}

\newpage
\input{./journal_docs/appendix.tex}
\end{document}

%% file: macros.tex
\definecolor{myred}{RGB}{153,51,51}
\definecolor{mygray}{RGB}{102,102,102}
\definecolor{darkgreen}{RGB}{60,160,60}
\definecolor{lightblue}{RGB}{60,60,200}
\definecolor{darkred}{RGB}{70,10,10}
\definecolor{darkblue}{RGB}{1,1,255}
\definecolor{yellow}{RGB}{255,255,0}
\definecolor{purple}{RGB}{127,0,255}
\definecolor{dark}{RGB}{1,1,1}
\definecolor{lightgray}{RGB}{230,230,230}
\definecolor{ucla_gold}{RGB}{255,232,0}
\definecolor{ucla_blue}{RGB}{50,132,191}
\definecolor{lightyellow}{rgb}{1.0, 1.0, 0.88}

\usepackage[capitalize]{cleveref}
\crefformat{equation}{(#2#1#3)}
\renewcommand{\Cref}[1]{\cref{#1}}

\newcommand{\tbf}[1]{\textbf{#1}}

\newcommand{\aed}[1]{\begin{aligned} #1 \end{aligned}}
\newcommand{\beq}[1]{\begin{equation}#1\end{equation}}

\newcommand{\rbr}[1]{\left(#1\right)}
\newcommand{\sbr}[1]{\left[#1\right]}
\newcommand{\cbr}[1]{\left\{#1\right\}}

\newcommand{\abs}[1]{\left|#1\right|}
\DeclarePairedDelimiter{\norm}{\lVert}{\rVert}

\providecommand\f[2]{\ensuremath \frac{#1}{#2}}

\DeclareMathOperator*{\argmin}{\text{argmin}}
\DeclareMathOperator*{\argmax}{\text{argmax}}

\DeclareMathOperator{\tr}{\text{Tr}}

\DeclareMathOperator*{\E}{\mathbb{E}}

\definecolor{color_skyblue}{rgb}{0.01,0.39,0.75}

\def \th {\theta}
\def \a {\alpha}
\def \b {\beta}

\def \l {\lambda}

\def \Om {\Omega}

\def \CC {\mathcal{C}}

\def \RR {\mathcal{R}}

\def \reals {\mathbb{R}}

\newcommand{\ignore}[1]{}

%% file: journal_docs/abstract.tex

\begin{abstract}
Domain shift, the mismatch between training and testing data characteristics,
causes significant degradation in the predictive performance in multi-source imaging scenarios.
In medical imaging, the heterogeneity of population, scanners and acquisition protocols at different sites presents a significant domain shift challenge and has limited the widespread clinical adoption of machine learning models. Harmonization methods, which aim to learn a representation of data invariant to these differences are the prevalent tools to address domain shift, but they typically result in degradation of predictive accuracy. This paper takes a different perspective of the problem: we embrace this disharmony in data and design a simple but effective framework for tackling domain shift. The key idea, based on our theoretical arguments, is to build a pretrained classifier on the source data and adapt this model to new data. The classifier can be fine-tuned for intra-study domain adaptation. We can also tackle situations where we do not have access to ground-truth labels on target data; we show how one can use auxiliary tasks for adaptation; these tasks employ covariates such as age, gender and race which are easy to obtain but nevertheless correlated to the main task. We demonstrate substantial improvements in both intra-study domain adaptation and inter-study domain generalization on large-scale real-world 3D brain MRI datasets for classifying Alzheimer’s disease and schizophrenia.\\

\keywords{Heterogeneity $\cdot$ Distribution shift $\cdot$ Domain adaptation $\cdot$ Domain generalization $\cdot$ MRI}

\end{abstract}

%% file: journal_docs/introduction.tex

\section{Introduction}

Deep learning models have shown great promise in several fields related to medicine, including medical imaging diagnostics~\citep{esteva2017dermatologist,rathore2017review} and predictive modeling~\citep{bashyam2020mri}.
Applications of medical imaging range from relatively common segmentation tasks~\citep{menze2014multimodal}, to more complex and high level decision-support functions, such as estimating different patterns of brain diseases~\citep{dong2015chimera,varol2017hydra,chand2020two} and producing personalized prognosis~\citep{rathore2018radiomic}.
However, despite their promise, complex deep learning models tend to have poor reproducibility across hospitals, scanners, and patient cohorts, since these high-dimensional models can overfit specific studies, and hence achieve modest generalization performance~\citep{davatzikos2019machine,zhou2020review}. While a potential solution to this weakness is to train on very large databases of diverse studies, this approach is limited in several ways.
Firstly, the characteristics of imaging devices change constantly, and hence even amply trained models are bound to face the same generalization challenges for new studies. Secondly, training labels, such as clinical or molecular classifications, or treatment measurements, are often scarce and hard to obtain.
It is therefore impractical to expect that such ample training is possible in many problems.
Finally, even if it were possible to train a model on large and diverse databases that would cover all possible variations across images, such a model would almost certainly sacrifice accuracy in favor of generalization under diverse conditions, i.e. it would have to rely on coarse imaging features that are stable across imaging devices and patient populations, and might fail to capture subtle and highly informative detail.

Herein, we propose a domain adaptation framework, which overcomes these limitations by allowing trained models to adapt to new imaging conditions in two paradigms: intra-study adaptation and inter-study generalization.
To improve the prediction accuracy of a model on heterogeneous images within each single study, intra-study adaptation strategy fast adapts a model which is pre-trained on the entire study to each sub-groups, e.g. age range, race, scanner type, by fine-tuning.
We use label information in the re-training process in this situation.
For adaptation between different studies, our inter-study generalization method can avoid using ground truth from the unseen study in view of the scarcity of labels in medical imaging.
Fundamental in our approach is the utilization of ``auxiliary tasks'', i.e., learning tasks that can be performed on readily available data from a new imaging condition (scanner, site, or population), and which can be used to adapt the primary trained model (e.g. disease classification) to these new conditions. An example of auxiliary tasks are estimation of readily available demographic characteristics, since such data is amply available in most practical settings.
Essentially, the auxiliary tasks help an already trained model adapt to new imaging conditions, by adapting the features extracted by  networks that are shared between the primary learning task and the auxiliary tasks.
We conducted extensive experiments on clinical large-scale studies of 2,614 3D T1-MRI scans to evaluate the effectiveness of the proposed framework for both Alzheimer’s disease and schizophrenia classification tasks.
Experimental results indicate that our proposed framework  substantially improves the performance in both intra-study adaptation and inter-study generalization paradigms.

\paragraph{Contributions}
Our main contributions are as follows.
\begin{itemize*}
  \item We discuss the necessity of adaptation instead of learning invariant representations for accurate prediction. We also introduce a regularization term in fine-tuning process of domain adaptation for diverse population and imaging devices.
  \item We propose a novel auxiliary task-based domain generalization method, that is able to adapt a model to an unseen study without accessing to prediction labels, with the guidance of easily accessible demographic information.
  \item We conduct extensive experiments on two classification tasks to evaluate the effectiveness of the proposed method. Our framework is superior to the baseline models according to the experimental comparison results.
\end{itemize*}

\paragraph{Organization of the manuscript}
We have organized this manuscript as follows. We first provide a detailed description of related work on domain adaptation and domain generalization in~\cref{sec:related_work}. Our goal is to compare and contrast existing methods and motivate our approach, which is described in detail in~\cref{sec:methods}. Next, we discuss the iSTAGING and PHENOM consortia, and details of the experimental setup in~\cref{sec:experiments}, followed by experimental results on intra-study and inter-study classification in~\cref{sec:results}. We provide a detailed discussion of these results along with pointers for future work in~\cref{sec:discussion}.

%% file: journal_docs/related_work.tex

\section{Related Work}
\label{sec:related_work}

We discuss the related literature in this section. We focus on the main techniques that have been shown to be suitable to handle domain-shift in the medical imaging and computer vision literature.


\paragraph{Non-deep-learning-based methods for harmonization}
Several non-deep learning-based methods for harmonization have been proposed to correct the bias in multi-site medical imaging data, e.g., the effect of the scanner. Methods based on parametric empirical Bayes~\citep{morris1983parametric} such as ComBat methods~\citep{johnson2007adjusting,pomponio2020harmonization} remove sources of variability, specifically site differences, by performing location and scale adjustments to the data. A linear model estimates the location and scale differences in cross-site data, while preserving other biologically-relevant covariates, such as sex and age. Methods such as~\citet{wachinger2021detect} explicitly add features pertaining to non-biological variability, e.g., scanner manufacturer's ID or magnetic field strength, into the model to tackle the fact that Combat algorithms may be insensitive to these variabilities.

\paragraph{Distribution alignment to learn invariant features}
Building representations of the source and target data that are invariant to their differences has be achieved using two main directions, (i) by employing discrepancy measures between two distributions and (ii) by using adversarial losses to build invariance. This suite of techniques are also called distribution alignment. Methods in first group use maximum mean discrepancy (MMD)~\citep{tzeng2014deep,long2015learning} or correlation distance~\citep{sun2016return} to measure distribution alignment. Adversarial adaptation methods such as~\citet{ganin2016domain,tzeng2017adversarial,liu2018detach,meng2020unsupervised,dinsdale2020unlearning} use a convolutional deep network to approximate the discrepancy. Domain classifier (discriminator) and domain-adversarial loss are used to learn features that are invariant across domains. The key issue with these methods is that the feature space of 3D MRI data is more complex than RGB data where these techniques have been primarily developed. This makes it difficult to measure distributional discrepancies and align the feature distribution for source and target data. Also, although adversarial adaptation techniques work well, optimizing adversarial objectives is very challenging and unstable in practice, especially for MRI data.

\paragraph{Learning disentangled representations}
An alternative to aligning the entire distribution of features is to disentangle the representation; in this case one learns two sets of features, the first which are specific to source or target data (also known as nuisances) and a second set of features (sufficient statistics) that are common to the two and thereby useful to build a robust classifier. Mutual information-based criteria are popular to disentangle the features. For instance,~\citet{meng2020learning} aims to extract generalized categorical features by explicitly disentangling categorical features and domain features via mutual information (MI) minimization. Since computing mutual information for real-valued features is very difficult, methods based on generative models such as~\citet{moyer2018invariant,moyer2020scanner} disentangle the latent representation using conditional variational autoencoders (VAEs)~\citep{kingma2013auto,sohn2015learning}. Related works such as~\citet{dewey2020disentangled} disentangle the latents by extracting high-resolution anatomical information and low-dimensional scanner-related components separately. A common trait and disadvantage of these methods is that they require access to data from multiple sources \emph{during training}. While this is reasonable for situations where agreements between different stake-holders make such sharing of data possible, it would be ideal if a method did not require data from all sources at the training time.

\paragraph{Domain translation methods}
To better model variations in the data, domain translation methods seek to learn image-to-image transformation between the source and target tasks. Recent works such as~\citet{hoffman2018cycada,bashyam2020medical,robey2020model,robinson2020image} utilize generative adversarial networks (GANs)~\citep{goodfellow2014generative,zhu2017unpaired,huang2018multimodal} to learn source-to-target domain mapping for images, where cycle-consistency is enforced for learning a  domain-invariant feature representation. Other works such as~\citet{shin2020gandalf} transfer the knowledge from MR images to positron emission tomography (PET) images by utilizing GANs for better diagnosis of Alzheimer’s disease. However, heterogeneity in data that comes from gender, age, ethnicity, and pathology, might not be preserved in an unpaired translation, especially when subjects are different in the source and target data. Furthermore, it is inefficient and ineffective to train multiple generative models in order to learn all mappings between multiple sites / domains. GANs, in particular for MRI data, are also notoriously difficult to train.

\paragraph{Building robust representations using techniques in causality}
Several recent studies incorporate causal inference~\citep{scholkopf2019causality} for learning robust representations. This is conceptually an extension of the idea of learning disentangled representations where one is interested in ensuring that the classifier only uses features that are causal predictors of the outcome. Causality-aware models~\citep{arjovsky2019invariant,heinze2021conditional} learn invariant features using regularizers on domain-specific information in the training data; this allows the representation to generalize to new domains. There are also other works such as~\citet{li2020domain} that extend this idea, or~\citet{zhang2020domain} which uses a graphical model to encode the domain variation and treats domain adaptation as Bayesian inference problem. Similarly~\citet{zhang2020causal} build a causal graph to generate perturbation (intervention) for data augmentation.
\cite{wang2021harmonization} construct a structural causal model with bijective mapping functions to modify the covariates, such as the scanner, for counterfactual inference.

\paragraph{Transfer learning and few-shot learning-based methods}
Few-shot learning methods seek to adapt a machine learning model to new classes with few labeled data by pretraining on other classes which may have abundant data available to train on. The broader problem is known as meta-learning or ``learning to learn''~\citep{thrun2012learning}. This has also found import in medical problems, e.g.,~\citet{qiu2020meta} utilize meta-learning to learn a new task with few samples in cancer survival analysis, while~\citet{dou2019domain} introduce global class alignment as well as local sample clustering regularization in a meta-learning scheme for domain generalization. A theme that has emerged in recent literature is that transfer learning, i.e., training a classifier on the abundant training data using standard classification losses and fine-tuning it to the new data, is an effective strategy to tackle few-shot problems~\citep{dhillon2019baseline,kolesnikov2019big}. Our methods for intra-study domain adaptation, where we pretrain a classifier on all available data but adapt it to data from a specific sub-group of the population, e.g., all people within a specific age-group, are directly motivated from the success of transfer and few-shot learning.

%% file: journal_docs/method.tex
\section{Methods}
\label{sec:methods}

This section gives the details of our technical approach. We will first introduce notation and concepts using intra-study domain generalization problems where we also provide theoretical arguments that are the foundation of our approach. We then elaborate upon our techniques for inter-study domain generalization.

\subsection{Problem Formulation}

Let $(x,y)$ denote the input datum and ground-truth labels, respectively. Labels $y \in C$ belong to a finite set. The training dataset consists of $N$ samples $D = \cbr{(x^i, y^i)_{i=1}^N}$ where each pair is drawn from the probability distribution $(x^i,y^i) \sim P$. A deep neural network is a machine learning model that predicts the probability of each output $y \in C$ given the input datum $x^i$ using parameters (weights) $\th \in \reals^n$. We denote the output of a deep network as a probability distribution over labels $p_\th(y \mid x)$. The goal of learning is to achieve good generalization, i.e., obtain accurate predictions over samples from the probability distribution $P$ that may not be a part of the training set $D$. The best weights for this purpose are
\beq{
    \th^* = \argmin_\th R(\th) := \E_{(x,y) \sim P} \sbr{-\log p_\th(y \mid x)},
    \label{eq:population_risk}
}
where the prediction of the model is $\hat{y}(\th) = \argmax_{y \in C} p_{\hat{\th}}(y \mid x)$.
The quantity $R(\th)$ is called the population risk of the model.
However, we only have access to samples $D$ from the distribution $P$ and therefore find the best weights
that fit a training dataset. This is achieved by minimizing the cross-entropy objective
\beq{
    \hat{\th} = \argmin_\th \f{1}{\abs{D}} \sum_{(x,y) \in D} -\log p_\th(y \mid x) + \Om(\th),
    \label{eq:meta_training}
}
where $\Om(\th)$ is a regularization term, e.g., $\Om(\th) = \l \norm{\th}_2^2/2$ for some constant $\l > 0$, that controls the amount of over-fitting on the training data. This objective is minimized using stochastic gradient descent~\citep{bottou2010large}.

\subsection{The Need for Adaptation}
If the training data is diverse, the training procedure above may not work well. To understand this, consider the case when the true distribution $P$ is a mixture of two sub-groups, i.e., $P = (P^{g_1} + P^{g_2})/2$ and similarly $D = D^{g_1} \cup D^{g_2}$. In the context of the present paper, these sub-groups may consist of data from subjects within a specific age range, gender, race or inputs with the same scanner type. We seek to understand how the model $\hat{\th}$ trained using~\cref{eq:meta_training} on the entire training data performs on one of the groups, say $g_2$, as compared to the best model trained only on $D^{g_2}$.
The development of~\citet{ben2010theory} shows that
with probability at least $1-\delta$ over independent draws of the training dataset from the distribution $P$,
\beq{
    R^{g_2}(\hat{\th}) \leq R^{g_2}({\th^*}^{g_2}) + c \sqrt{\f{V - \log \delta}{N}} + \f{d(P^{g_1}, P^{g_2})}{2} + R(\th^*);
    \label{eq:domain_adaptation_bound}
}
here $c$ is a constant, $R^{g_2}(\hat{\th})$ is the population risk on group $g_2$ using our learned weights $\hat{\th}$ using the training data and $R^{g_2}({\th^*}^{g_2})$ is the best population risk on $g_2$ that one could have obtained by using~\cref{eq:population_risk} using data only from distribution $P^{g_2}$. The constant $V$ is the Vapnik-Chervonenkis dimension~\citep{blumer1989learnability} that characterizes the complexity of a deep network architecture. The third term $d(P^{g_1}, P^{g_2})$ is a measure of the diversity of data from the two sub-groups and the final term is the best population risk using data from both groups.
This inequality, and an analogous expression for group $g_1$, is particularly illuminating. Note that we want the population risk on group $g_2$, i.e., the left-hand side $R^{g_2}(\hat{\theta})$ using our chosen weights $\hat{\theta}$ to be as close as possible to the best population risk $R^{g_2}({\theta^*}^{g_2})$ for that sub-group. Ideally, the last three terms on the right-hand side should be small. First, if $R(\th^*)$ is large, i.e., there does not exist a well-performing model of our chosen architecture that can fit both sub-groups, then we expect the learned model $\hat{\th}$ to also work poorly on the group $g_2$. This directly suggests that one must use a deep network with large learning capacity if the sub-groups are diverse. Second, larger the deep network, larger the capacity $V$ and more the data $N$ necessary to achieve a good generalization. Third, the term $d(P^{g_1}, P^{g_2})$ suggests that fixed the capacity $V$ and number of data $N$, if data from the two sub-groups is diverse then the learned model $\hat{\th}$ may not be accurate on any of the sub-groups.

In practice, data from different sub-groups such as age range, gender, race, and scanner type, can be quite diverse. The above discussion suggests that it is difficult to learn machine learning models that generalize well for each sub-group using data from the entire heterogeneous population. This is a fundamental hurdle to building performant machine learning models for clinical applications and we next propose a simple solution to this problem.

\subsection{Intra-Study Domain Generalization}

Consider the situation when we wish to obtain accurate predictions on one sub-group of the population, corresponding to the subset of the training data $D^g \subset D$. A naive way of doing so is to simply sequester the training dataset and employ~\cref{eq:meta_training} to learn a model $\hat{\th}^g$ using only data $D^g$. If we have $m$ mutually exclusive sub-groups in the training data $D = D^{g_1} \cup D^{g_2} \ldots \cup D^{g_m}$, all say with equal amounts data $N/m$, then a classical bound~\citep{wellner2013weak} on the generalization performance for each sub-group of this naive sequestering is given by
\beq{
    R^g(\hat{\theta}^g) \leq R^g({\theta^*}^g) + c \sqrt{\f{m(V - \log \delta)}{N}};
    \label{eq:vc_bound_one_group}
}
notice that the second term has degraded by a factor of $\sqrt{m}$ as compared to the situation if there were only one sub-group in the data. This degradation occurs because the model $\hat{\theta}^g$ is fitted only on data from sub-group $g$.
In this paper, we avoid this degradation by a simple modification to~\cref{eq:meta_training}. Roughly speaking, our goal is to fit a classifier $D^g$ on a restricted class of classifiers, namely the ones that are close to the classifier $\hat{\theta}$ trained on the entire training dataset and reduce the factor of $\sqrt{m}$ above.
To that end, we solve the optimization problem given by
\def \thgdr {\hat{\th}^g_{\text{DR}}}
\beq{
    \thgdr = \argmin_{\th} \f{1}{\abs{D^g}} \sum_{(x,y) \in D^g} -\log p_\th(y \mid x) + \f{\a}{2} \norm{\th - \hat{\th}}^2_2,
    \label{eq:fine_tuning}
}
where $\hat{\theta}$ is the model trained on the entire study $D$ using~\cref{eq:meta_training}. The objective is quite similar to the cross-entropy objective in~\cref{eq:meta_training} except that we have included an additional term, which we call a ``proximal'' term, that depends upon the hyper-parameter $\a > 0$. This term encourages the new weights $\hat{\th}^g$ to be close to the original learned weights $\hat{\th}$. Roughly speaking, if $\a$ is large, the optimization problem keeps the weights close to the pre-trained weights $\hat{\theta}$. It is therefore beneficial to pick a large value of $\alpha$ if the number of samples in the subset $D^g$ is small, or $D^g$ is too different from the rest of the training data in $D$. A small value of $\alpha$ is ideal for the complementary cases, namely if $D^g$ has a large number of samples which enables fitting a low-variance classifier in~\cref{eq:fine_tuning}, or if the sub-group $D^g$ has data similar to the other sub-groups.

\paragraph{The technical argument for intra-study domain generalization for a restricted class of classifiers}
We next present a mathematically precise argument for our intra-study domain generalization methodology using the doubly robust estimation framework~\citep{reddi2015doubly}. We will work in the restricted setting of a kernel-based binary classifier, i.e., $C = \cbr{-1,1}$, denoted by $f: X \mapsto Y$, that maps the inputs $x \in X$ to their labels $y \in Y$. In the kernel setting we can write $f(x) = \sum_{i=1}^N \theta_i k(x^i, x)$ where $k(\cdot, \cdot)$ is called the ``kernel'' that measures the similarity between a new datum $x$ and datum $x^i$ from the training set. The parameters $\theta_i$ parametrize the classifier in terms of these similarities and are conceptually similar to the weights of a neural network. In this case, a bound on the Vapnik-Chervonenkis dimension $V$ in~\cref{eq:vc_bound_one_group} can be explicitly computed and the population risk which we denote by $R(f)$ is a convex function of the classifier $f$. As done in~\cref{eq:meta_training}, let $\hat{f}$ be the classifier
\beq{
    \hat{f} = \argmin_{f:\ \Om(f) \leq \nu} \f{1}{n} \sum_{i=1}^n \ell(f, x^i, y^i);
    \label{eq:kernel_meta_training}
}
that minimizes the average misprediction loss $\ell(f, x^i, y^i)$ over all the training data. We have written the regularization $\Om(\theta)$ in~\cref{eq:meta_training} slightly differently here. We have written the training as a constrained minimization problem with the constraint $\Om(f) \leq \nu$; the two versions are equivalent to each other. In the kernel setting, the regularization akin to $\Om(\theta) = \l \norm{\theta}_2^2/2$ employed in~\cref{eq:meta_training} is written as $\Om(f) = \l \norm{f}^2/2 = \f{1}{2} \sum_{i,j} \theta_i \theta_j k(x^i, x^j)$ where the norm is an appropriate function-space norm (the so-called Reproducing Kernel Hilbert Space (RKHS) norm). The error bound from~\cref{eq:vc_bound_one_group} now looks like
\beq{
    R^g(\hat{f}^g) \leq R^g({f^*}^g) + c \sqrt{\f{m \rbr{\nu^2 \tr(K) - \log \delta}}{N}},
    \label{eq:kernel_bound}
}
with probability at least $1-\delta$ over draws of the dataset~\citep{bartlett2002rademacher}; here $c > 0$ is a constant, $\tr(K)$ is the trace of a certain kernel matrix of the training dataset and $f^*$ is the classifier that minimizes the population risk on the distribution $P$.

If we now perform intra-study domain generalization to fit data from sub-group $g$, we should compute
\beq{
    \hat{f}^g_{\text{DR}} = \argmin_{f: \Om(f-\hat{f}) \leq \nu_{\text{DR}}} \f{1}{\abs{D^g}} \sum_{(x,y) \in D^g} \ell(f, x, y);
    \label{eq:kernel_dr_objective}
}
the important thing to note here is that we are restricting the solution to be in the neighborhood of the pre-trained classifier $\hat{f}$ using the regularization $\Om(f-\hat{f}) \leq \nu_{\text{DR}}$. The result from~\citet[Theorem 4]{reddi2015doubly} then gives the following: with probability at least $1-\delta$ over different draws of the training dataset,
\beq{
    R^g(\hat{f}^g_{\text{DR}}) \leq R^g({f^*}^g)
    + c \sqrt{\f{{\nu'}^2 \tr(K) - \log \delta}{N}} + \f{c' \norm{\hat{f}}}{N}.
    \label{eq:kernel_doubly_robust_bound}
}
where $\nu' = \nu_{\text{DR}} + \sqrt{\f{\nu^2 \tr(K) - \log \delta}{N}}$. The most important aspect of the above inequality, that directly motivates our approach in this paper, is that we have traded off the multiplicative factor of $\sqrt{m}$ in~\cref{eq:kernel_bound} with the additive term proportional to $\f{ \norm{\hat{f}}}{N}$.
If the norm of the pretrained classifier $\norm{\hat{f}}$ is small (conceptually this means that the hypothesis $\hat{f}$ is a ``simple'' function) then the right-hand side of~\cref{eq:kernel_doubly_robust_bound} can be much smaller than that of~\cref{eq:kernel_bound}. There are recent theoretical results that suggest that even non-kernel-based classifiers such as deep networks when trained with stochastic optimization algorithms result in simple hypotheses~\citep{belkin2019reconciling}. Roughly speaking, we have traded the variance in the classifier caused by the small sample size in~\cref{eq:kernel_bound} for the bias in~\cref{eq:kernel_doubly_robust_bound} that arises from restricting the class of functions that are fitted to the new sub-group. Doing so is likely to be beneficial if the sub-group $g$ has few samples.

\subsection{Inter-Study Domain Generalization}
\label{sec:inter-study}

In this section, we extend our arguments to inter-study domain generalization. Our approach builds upon intra-study generalization described in the previous section. In simple words, we would like to pre-train a classifier on the source study using~\cref{eq:meta_training} and then adapt (or fine-tune) it to data from the target study~\cref{eq:fine_tuning}. The crucial difference however is that we may not have access to the ground-truth labels of the data from the target study and therefore cannot directly employ~\cref{eq:fine_tuning}. We must therefore adapt the pre-trained classifier using some other means, which leads us to our main innovation for inter-study domain generalization.

\paragraph{Using auxiliary tasks for domain generalization}
Let us denote the training data for the source study by $D^s = \cbr{(x^i, y^i, y_a^i)}_{i=1}^{N^s}$. We assume that we have access to two kinds of ground-truth labels, the first $y^i$ are the labels for the primary task, e.g., predicting the cognitive normal versus Alzheimer's disease. The labels $y_a^i$ for $a \in A$ consist of auxiliary attributes such as age, gender and race. The ideal auxiliary attribute is something that is readily available on both the source and the target study, and is correlated with the primary task label. Given target study data $D^t = \cbr{(x^i, y^i, y_a^i)}_{i=1}^{N^t}$, our domain generalization framework from the previous section can be now employed as follows.

\def \thb {\th_{\text{base}}}
\def \pmain {\varphi_{\text{main}}}
\def \paux {\varphi_{\text{aux}}}
\def \hthb {\hat{\th}_{\text{base}}}
\def \hhthb {\hat{\hat{\th}}_{\text{base}}}
\def \hpmain {\hat{\varphi}_{\text{main}}}
\def \hhpmain {\hat{\hat{\varphi}}_{\text{main}}}
\def \hhpmain {\hat{\hat{\varphi}}_{\text{main}}}
\def \hpaux {\hat{\varphi}_{\text{aux}}}

\begin{figure}[htpb]
\centering
\subfloat{\includegraphics[width=0.9\textwidth]{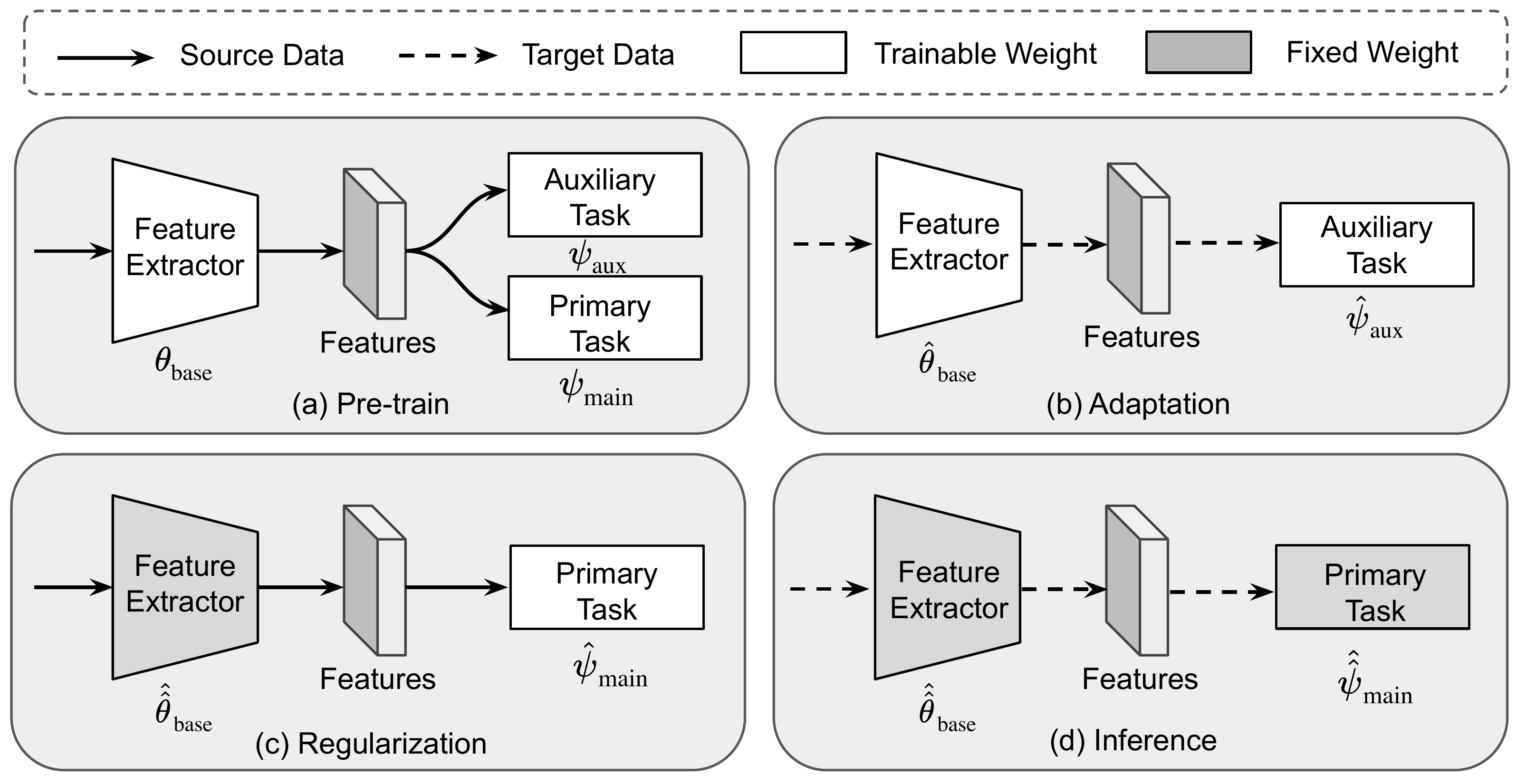}}
\caption{\textbf{Inter-study domain generalization framework.} There are four sequential phases in the framework: (a) pre-train, (b) adaptation, (c) regularization, and (d) inference. We use source-study data $D^s$ in phase (a) and (c); target-study data $D^t$ in phase (b) and (d). The source and target data flows are denoted as solid and dashed lines. White and grey blocks indicate trainable network and frozen network separately. We have denoted the weights of the various blocks in the picture, which are computed using~\cref{eq:step_1,eq:step_2,eq:step_3}.}
\label{fig:network}
\end{figure}

We build a deep network with the architecture depicted in~\cref{fig:network}. It consists of three parts, a feature extractor whose weights we denote by $\thb$, a multi-layer neural network-based primary (main) classifier with weights $\pmain$ which takes these features as inputs and predicts the primary task labels and another auxiliary classifier with weights $\paux$ which predicts all auxiliary labels using these features. Training these three networks proceeds in three steps discussed below.

\paragraph{Step 1: Pretraining} We pretrain a classifier on the source study to predict both labels of the primary task and the auxiliary labels. This involves solving
\beq{
    \aed{
    \hthb, \hpmain, \hpaux
    = \argmin \f{1}{\abs{D^s} \abs{A}} \sum_{a \in A} \sum_{(x,y,y_a) \in D} & -\log p_{\thb, \pmain}(y \mid x)
    + \b_a \ell_a(\thb, \paux, x, y_a)\\
    &+ \Om(\thb, \pmain, \paux).
    }
    \label{eq:step_1}
}
We use multiple auxiliary tasks, each with its own specific mis-prediction loss $\ell_a$ and coefficients $\b_a$ that are hyper-parameters chosen via cross-validation. This is step is analogous to building the classifier $\hat{f}$ in~\cref{eq:kernel_meta_training} using data from the entire training set.

\paragraph{Step 2: Adaptation of the feature extractor} We next fine-tune the pretrained feature extractor and the classifier for the auxiliary tasks using data from the \emph{target study}.
\beq{
    \hhthb, \hpaux = \argmin_{\thb, \paux} \f{1}{\abs{D^t} \abs{A}} \sum_{a \in A} \sum_{(x,y_a) \in D^t} \ell_a(\thb, \paux, x, y_a) + \f{\a}{2} \norm{\thb - \hthb}_2^2.
    \label{eq:step_2}
}
This step is a key innovation of our approach: since labels for auxiliary tasks such as age, gender and race are easily available, we can adapt the feature extractor to learn features that better suited to making predictions on the target data. Note that we do not perform any regularization on the weights of the auxiliary classifier, this enables large changes to the auxiliary classifier to predict the target data accurately. This step is similar to the fine-tuning step in~\cref{eq:fine_tuning} or its kernel version~\cref{eq:kernel_dr_objective} but instead of using a proximal penalty on both $\thb$ and $\paux$, we use the penalty only on $\thb$ to let the auxiliary classifier fit to the target data without any constraints. The rationale of doing so is to let the auxiliary classifier capture the variability between the source and target data through the auxiliary tasks without changing the feature extractor significantly.

\paragraph{Step 3: Adaptation of the primary task classifier}
With the feature extractor fixed to the its value $\hhthb$ from Step 2, we fine-tune the primary classifier on the source data to get
\beq{
    \hhpmain = \argmin_{\pmain} \f{1}{\abs{D^s}} \sum_{(x,y) \in D^s} - \log p_{\hhthb, \pmain}(y \mid x) + \Om(\pmain - \hpmain).
    \label{eq:step_3}
}
This step adapts the primary classifier to the modified features generated by $\hhthb$. Its goal is to obtain the primary classifier $\hhpmain$ that can accurately classify the primary task on the \emph{source} study using the modified features provided by the feature extractor $\hhthb$. Conceptually, this is nothing but the pretraining phase in~\cref{eq:meta_training} or~\cref{eq:kernel_meta_training} except that weights of the feature extractor do not change and weights of the primary classifier are the only ones that change.

\paragraph{Step 4: Inference on new data} from the target task is performed using weights of the feature extractor $\hhthb$ and weights of the primary classifier $\hhpmain$.

\begin{figure}[!t]
\centering
\subfloat{\includegraphics[width=0.75\textwidth]{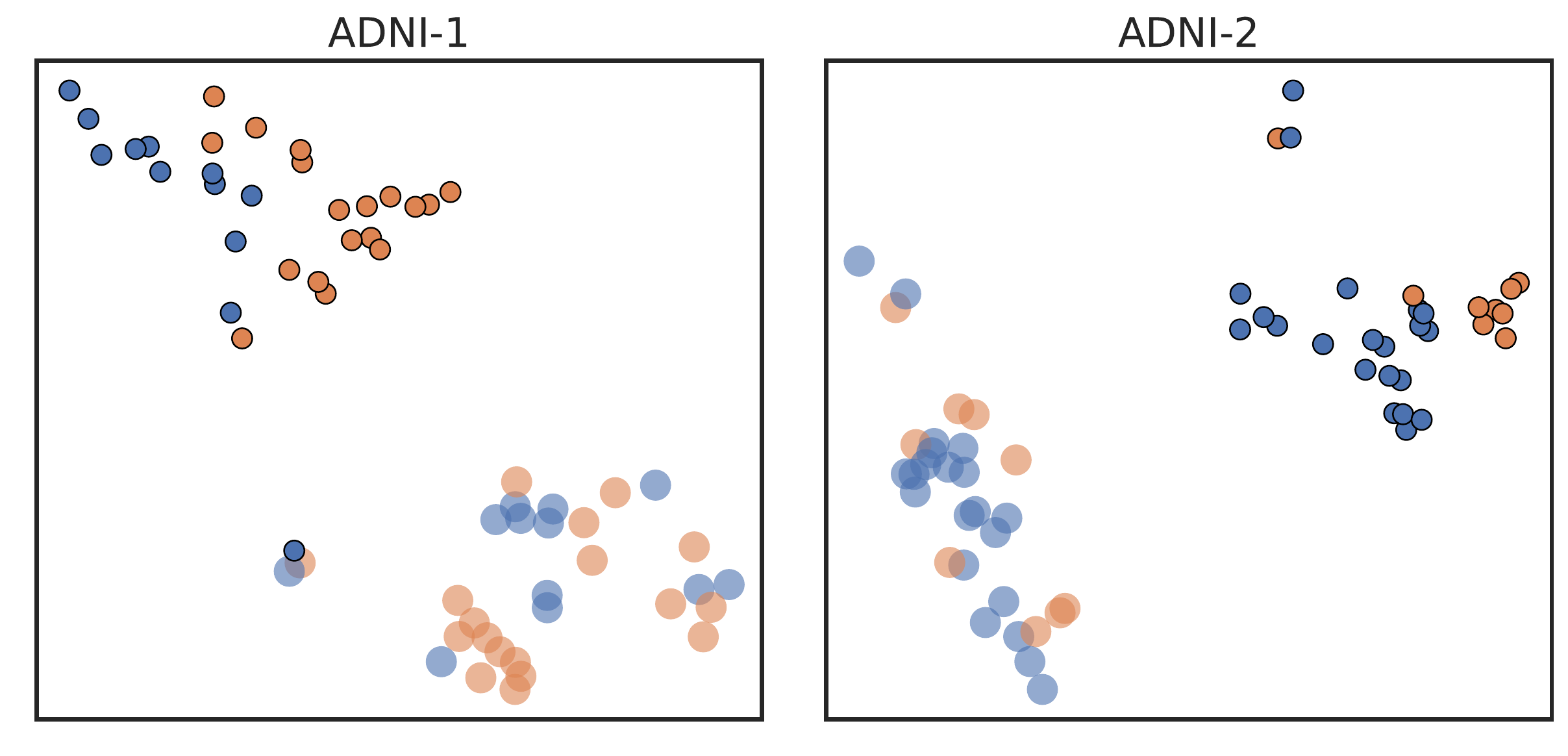}}
\caption{\textbf{t-SNE~\citep{van2008visualizing} embeddings of the features for ADNI-1 and ADNI-2 studies from a model trained on PENN}. Colors denote the ground-truth labels. Translucent markers and opaque markers denote the embeddings before and after adaptation respectively.
The figures show 2D t-SNE embeddings mapped from high-dimensional features that are extracted from the feature extractor. We observe that samples (translucent markers) from two two categories are entangled with each other before adaptation. After performing auxiliary task based adaptation, the samples (opaque markers) are well separated to two groups.}
\label{fig:tsne}
\end{figure}


%% file: journal_docs/experiments.tex

\section{Experiments}
\label{sec:experiments}

\subsection{Materials}

\begin{table}[!t]
\caption{\textbf{Summary of participant demographics in iSTAGING consortium.} CN and AD denote cognitive normal and Alzheimer's disease respectively. Age is described in format: mean $\pm$ std [min, max]. F and M in gender represent female and male separately. W, AA and A in race represent white, African American, and Asian separately.}
\label{tab:istaging}
\begin{center}
\begin{small}
\begingroup
\setlength{\tabcolsep}{4pt}
\begin{tabular}{l cccccc }
\toprule
Study & Subjects & CN & AD & Age & Gender (F/M) & Race (W/AA/A)  \\
\midrule
ADNI-1 & 369 & 178 & 191 & 75.5 $\pm$ 6.5 [55.0, 90.9]  & 178 / 191 & 341 / 21 / 5 \\
ADNI-2/GO & 407 & 261 & 146 & 73.1 $\pm$ 6.8 [55.4, 90.3] & 206 / 201 & 369 / 21 / 9 \\
ADNI-3 & 27 & 24 & 3 & 71.0 $\pm$ 7.1 [55.8, 89.2] & 18 / 9 & 27 / 0 / 0 \\
PENN & 572 & 229 & 343 & 72.0 $\pm$ 11.9 [22.6, 95.2] & 361 / 211 & 432 / 116 / 10 \\
AIBL & 568 & 481 & 87 & - & - & - \\
\bottomrule
\end{tabular}
\endgroup
\end{small}
\end{center}
\end{table}

\begin{table}[!t]
\caption{\textbf{Summary of participant demographics in PHENOM consortium.} NC and SCZ denote normal control and Schizophrenia respectively. Age is described in format: mean $\pm$ std [min, max]. F and M in gender represent female and male separately.}
\label{tab:phenom}
\begin{center}
\begin{small}
\begin{tabular}{l cccccc }
\toprule
Study & Subjects & NC & SCZ & Age & Gender (F/M) & Field Strength  \\
\midrule
Penn & 227 & 131 & 96 & 30.5 $\pm$ 7.2 [16.2, 45.0]  & 121 / 106 & 3.0T\\
China & 142 & 76 & 66 & 31.2 $\pm$ 7.5 [17.0, 45.0] & 69 / 73 & 3.0T \\
Munich & 302 & 157 & 145 & 29.4 $\pm$ 6.9 [18.0, 45.0] & 79 / 223 & 1.5T \\
\bottomrule
\end{tabular}
\end{small}
\end{center}
\end{table}

We validate our proposed method on 2,614 3D T1-weighted brain magnetic
resonance imaging (MRI) scans from two large-scale imaging consortia:
iSTAGING~\citep{habes2021brain} and
PEHNOM~\citep{satterthwaite2010association,wolf2014amotivation,zhang2015heterogeneity,zhu2016neural,zhuo2016schizophrenia,rozycki2018multisite,chand2020two}.
In the iSTAGING consortium, we investigate data from three sites, including
the Alzheimer’s Disease Neuroimaging Initiative
(ADNI)~\citep{jack2008alzheimer}, Penn Memory Center (PENN), and the
Australian Imaging, Biomarkers and Lifestyle
(AIBL)~\citep{ellis2010addressing}. Cognitive normal (CN) and  Alzheimer’s
disease (AD) are the two diagnosis groups that considered in this dataset. The
detailed description and clinical variables including age, gender, and race,
are shown in~\cref{tab:istaging}. In the PHENOM consortium, we use data
from multiple centers, including Penn, China, and Munich. There are two
diagnosis groups: normal control (NC) and Schizophrenia (SCZ) patients. A
detailed description of clinical variables including age and gender
is shown in~\cref{tab:phenom}. We also show site differences in terms of
scanner type, magnetic strength, and acquisition parameters in~\cref{site}.
For image preprocessing, we use a minimal pipeline since
neural networks can extract rich features automatically.
All images are first bias-field corrected with N4ITK~\citep{tustison2010n4itk},
and then aligned to a standard MNI~\citep{fonov2009unbiased} space using
ANTs~\citep{avants2008symmetric}. Registered images have $193\times 229
\times 193$ with 1 $\text{mm}^3$ isotropic voxels. We perform intensity
normalization on the data before feeding them into the network.

\subsection{Network and Implementations}

Our deep networks are designed to be small to improve memory requirements and computational efficiency; it is based on the architecture of~\citet{wen2020convolutional}. The feature extractor consists of five blocks where each block consists of one convolution layer, batch-normalization~\citep{ioffe2015batch}, rectified linear units (ReLU) nonlinearity, and max pooling. We enlarge the receptive field by using a $5 \times 5 \times 5$ convolutional kernel. A three-layer multilayer perceptron (MLP) with ReLU nonlinearity is used as the classifier for the primary task; the classifier for the auxiliary task is the same. More details of the architecture are provided in~\cref{arch}. All models are implemented using PyTorch~\citep{paszke2019pytorch}.

\paragraph{Hyper-parameters}
We are keenly interested in developing robust methods that perform well across a wide variety of evaluation benchmarks. \emph{Our architectures and hyper-parameters are consistent across all experiments.} We use Adam~\citep{kingma2014adam} with initial learning rate $10^{-4}$ and weight decay $10^{-5}$ for optimization. A learning rate scheduler with cyclic cosine annealing~\citep{he2019bag} is used for better convergence. We use batch size of 6 and train for 60 epochs. During training, we augment the input data using a Gaussian blur filter. For fine-tuning in intra-study domain adaptation experiments, we use $\lambda = 0.1$ and $\alpha = 0.01$. For inter-study experiments, we choose $\beta_a=1.0$ for cross-entropy loss in auxiliary task, and $\beta_a=0.1$ for those who use Huber loss as loss function.
We demonstrate a case study on hyper-parameter search in~\cref{subsec:hyperparam}.

\paragraph{Baselines and Experimental Setup}
We compare the performance of our proposed method with state-of-the-art algorithms such as Invariant Risk Minimization (IRM)~\citep{arjovsky2019invariant}, Adversarial Discriminative Domain Adaptation (ADDA)~\citep{tzeng2017adversarial}, Domain-Adversarial Neural Network (DANN)~\citep{ganin2016domain}, and cross-modal transformer (Fusion)~\citep{yang2021learning} for both AD classification and SCZ classification tasks as shown in~\cref{tab:istaging_result} and~\cref{tab:phenom_result}. IRM, ADDA, and DANN learn domain-invariant features by different techniques, e.g., causality-aware model for IRM and distribution alignment for ADDA and DANN. Fusion learns feature embedding from multi-modal data using the self-attention mechanism in Transformer models in natural language processing.
We used publicly available code by the original authors of these methods. These methods, in particular ADDA and DANN had extremely poor performance out of the box on our datasets and we therefore spent significant amounts of effort and time to search for hyper-parameters for the existing methods to be able to provide a fair comparison against them.
In our setup, we use sex and age information as the additional modality besides MRI images as inputs to the model. TarOnly and SrcOnly serve as two baselines that are trained directly on the target site and trained only on the source study and tested on the target study separately. We can think of the accuracy of TarOnly as an upper-bound on the target-study classification accuracy because it is the only algorithm that is trained directly on the inputs and labels from the target.

%% file: journal_docs/results.tex
\section{Results}
\label{sec:results}

\begin{figure}[!t]
\centering
\subfloat[Age groups in ADNI]{\includegraphics[width=0.49\textwidth]{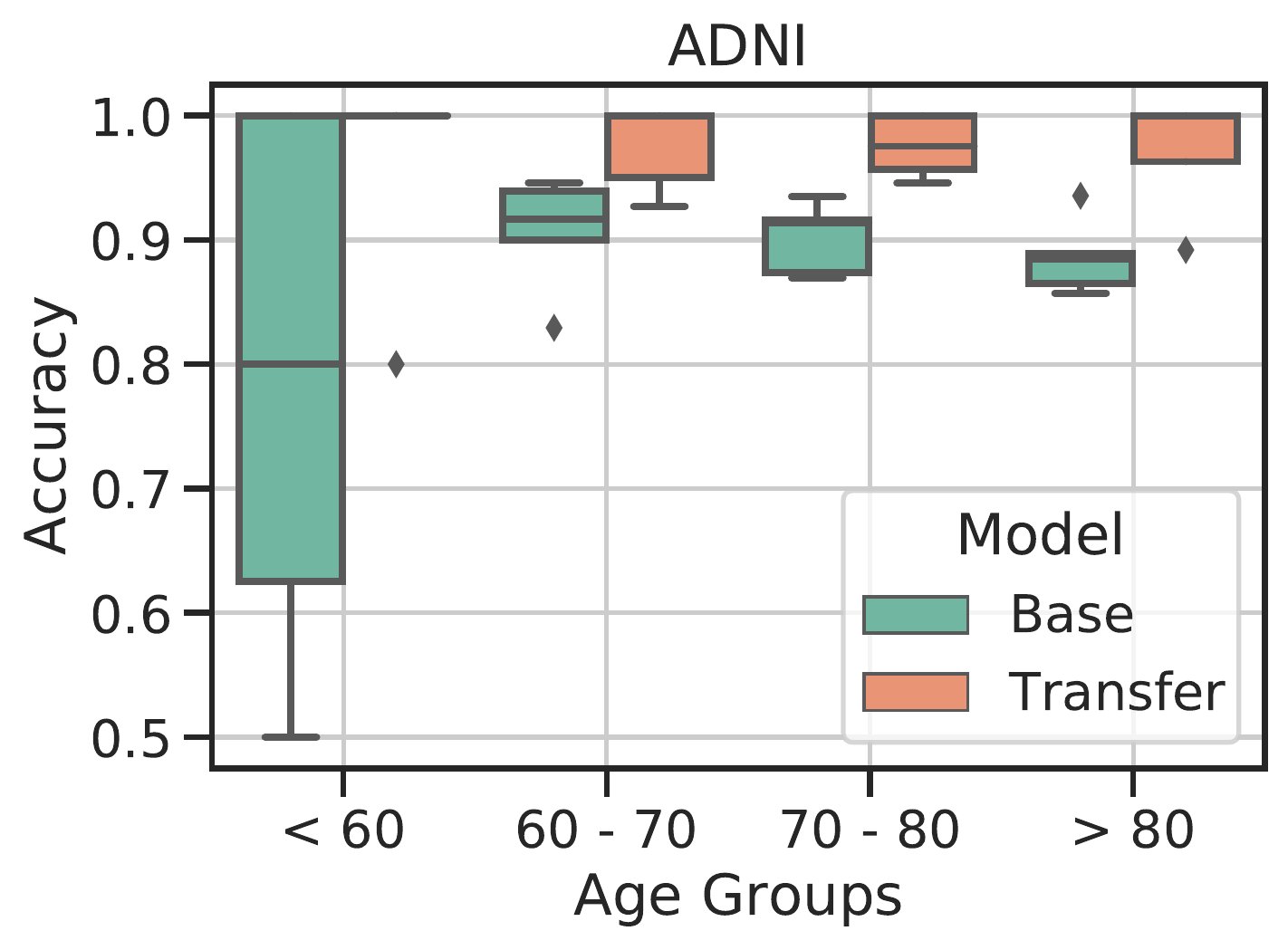}}
\subfloat[Race groups in PENN]{\includegraphics[width=0.5\textwidth]{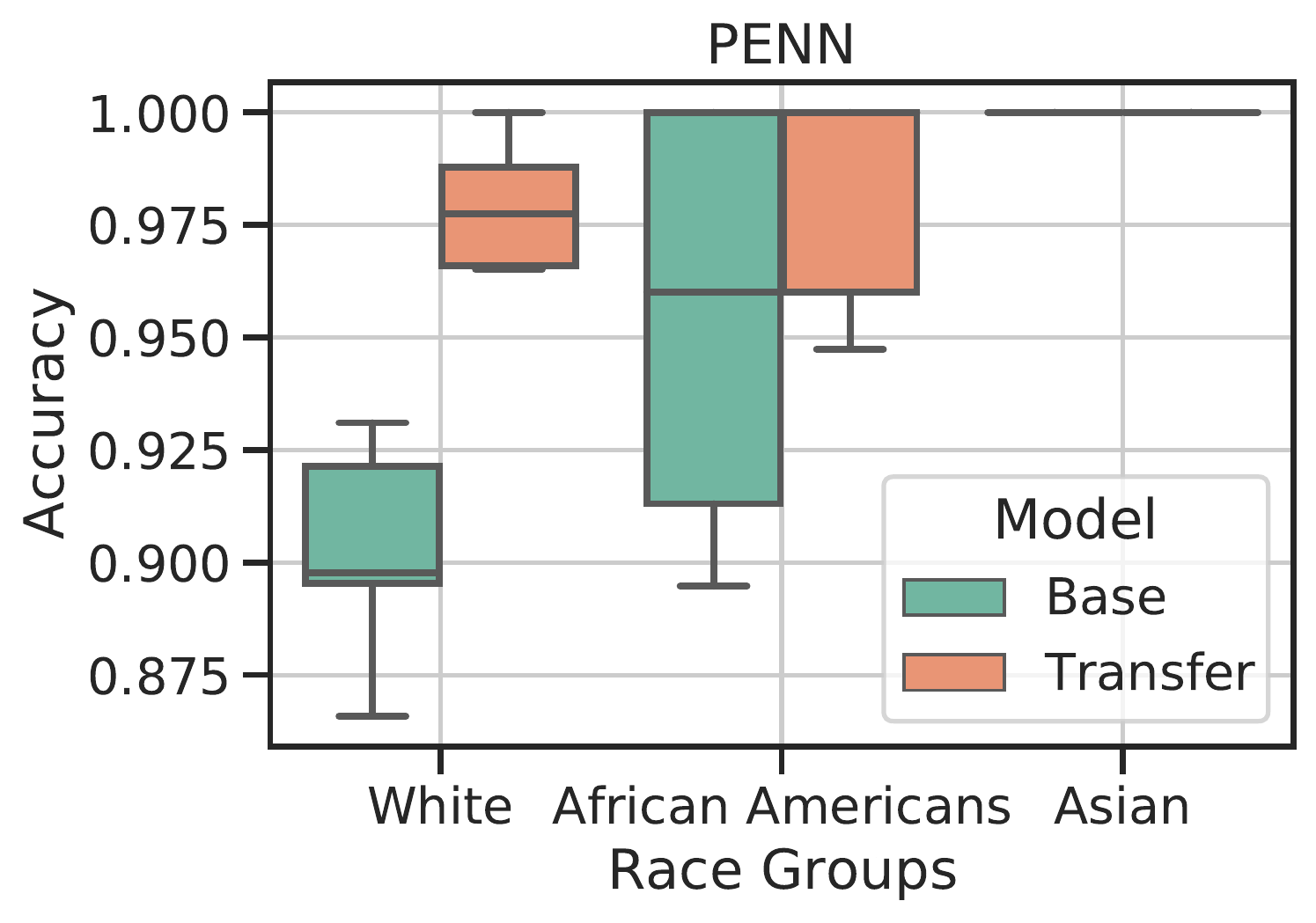}}\\
\subfloat[Scanner types in AIBL]{\includegraphics[width=0.5\textwidth]{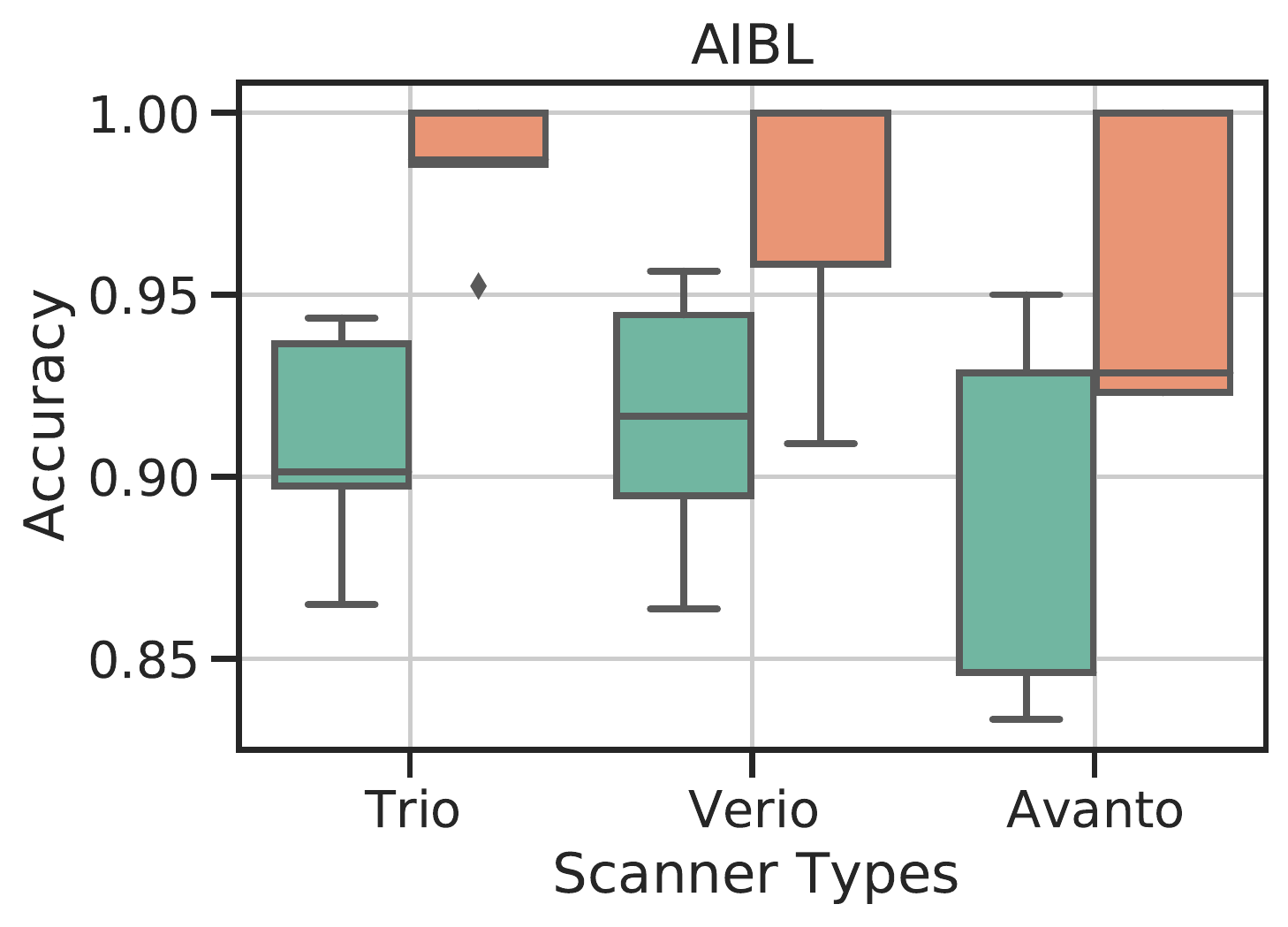}}
\caption{\textbf{Intra-study Alzheimer’s disease classification accuracy on domain-specific groups in iSTAGING consortium.} The transfer (adapted) models in each sub-group achieve higher accuracies and reduced variance compared to the base models in all studies.}
\label{fig:exp_single}
\end{figure}

\subsection{Intra-Study Alzheimer’s Disease Classification}

We perform Alzheimer's disease (AD) classification on the ADNI, PENN, and AIBL
studies from iSTAGING consortium separately. First, we train a base model on
all available data for each study individually and then evaluate each base
model using its classification accuracy across five-fold cross-validation.
Next, we fine-tune the models by training on data from each domain-specific
group to boost the classification accuracy. In ADNI, we fine-tune the base
model to four age groups: 3.4\% participants with age less than 60, 25.8\%
participants with age between 60 and 70, 51.5\% participants with age between
70 and 80, and 19.3\% participants with age greater than 80. As shown
in~\cref{fig:exp_single} (a), the AD classification of each transfer model
increased significantly compared to the corresponding base models in all age
groups. Similarly, we split data in PENN study into three race groups: 77.4\%
Caucasians, 20.8\% African American, and 1.8\% Asian. We find that the mean
classification accuracy increases and the variance is reduced for both
Caucasians and African Americans in~\cref{fig:exp_single} (b). Since there are
only 10 Asian subjects, we couldn't observe any substantiative improvement in
this group. For the AIBL study, images are collected from three types of
Siemens scanner including 65.0\% from Trio, 18.5\% from Verio, 16.5\% from
Avanto. In~\cref{fig:exp_single} (c), we also observe a substantial
improvement for all scanner subgroups. The domain adaptation approach can help
in providing more precise and accurate predictions compared to the base model
when domain-related information, such as age, race and scanner are available.

\begin{figure}[!tp]
\centering
{
\subfloat[Train on ADNI-1 and test on ADNI-2 and PENN.]{\includegraphics[width=0.9\textwidth]{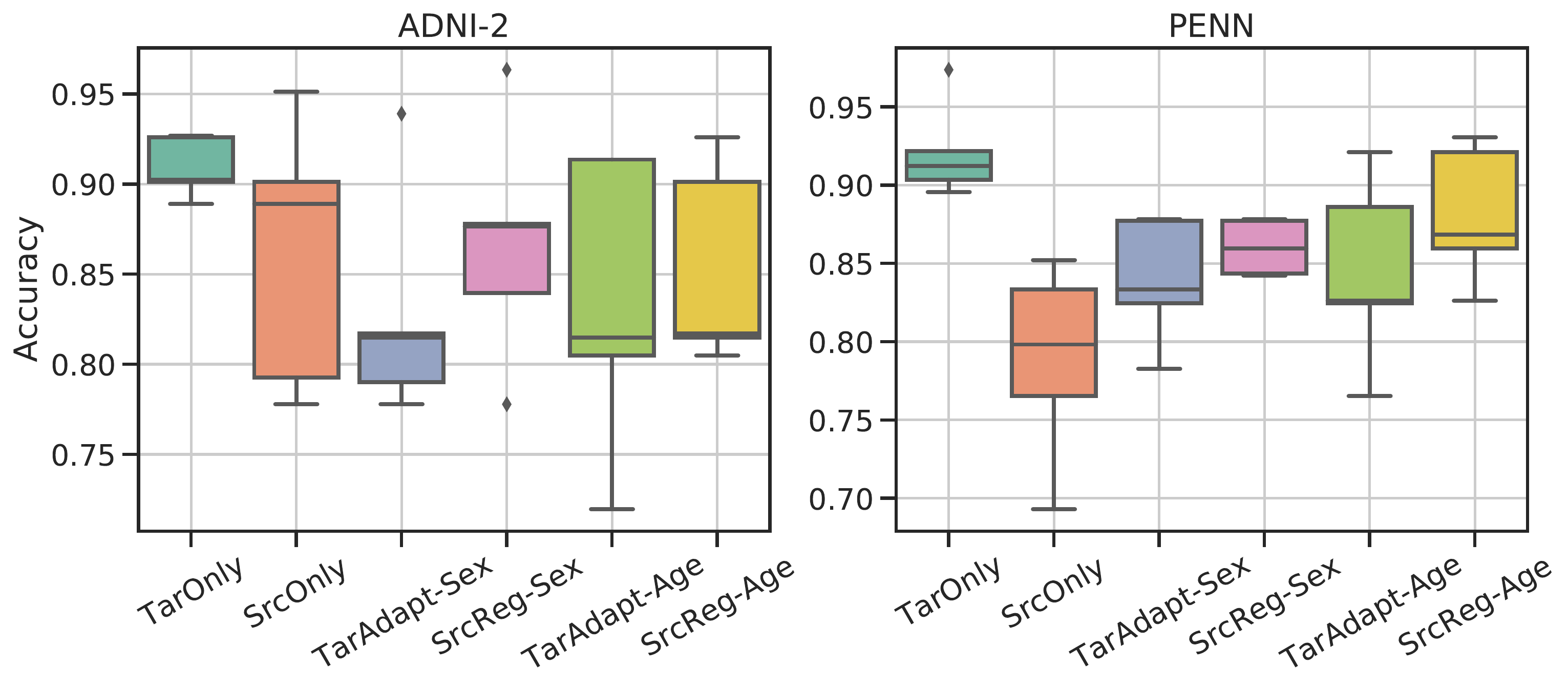}}\\
\subfloat[Train on ADNI-2 and test on ADNI-1 and PENN.]{\includegraphics[width=0.9\textwidth]{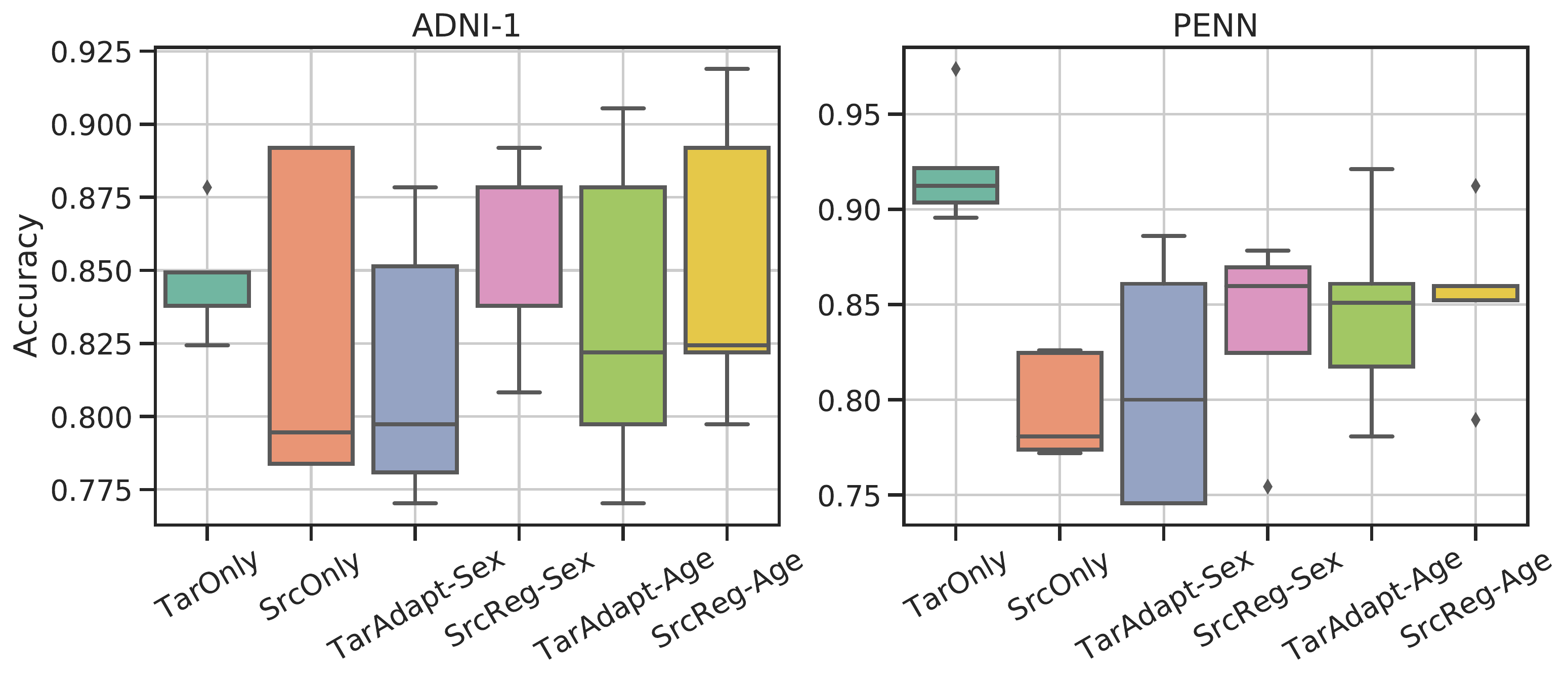}}\\
\subfloat[Train on PENN and test on ADNI-1 and ADNI-2.]{\includegraphics[width=0.9\textwidth]{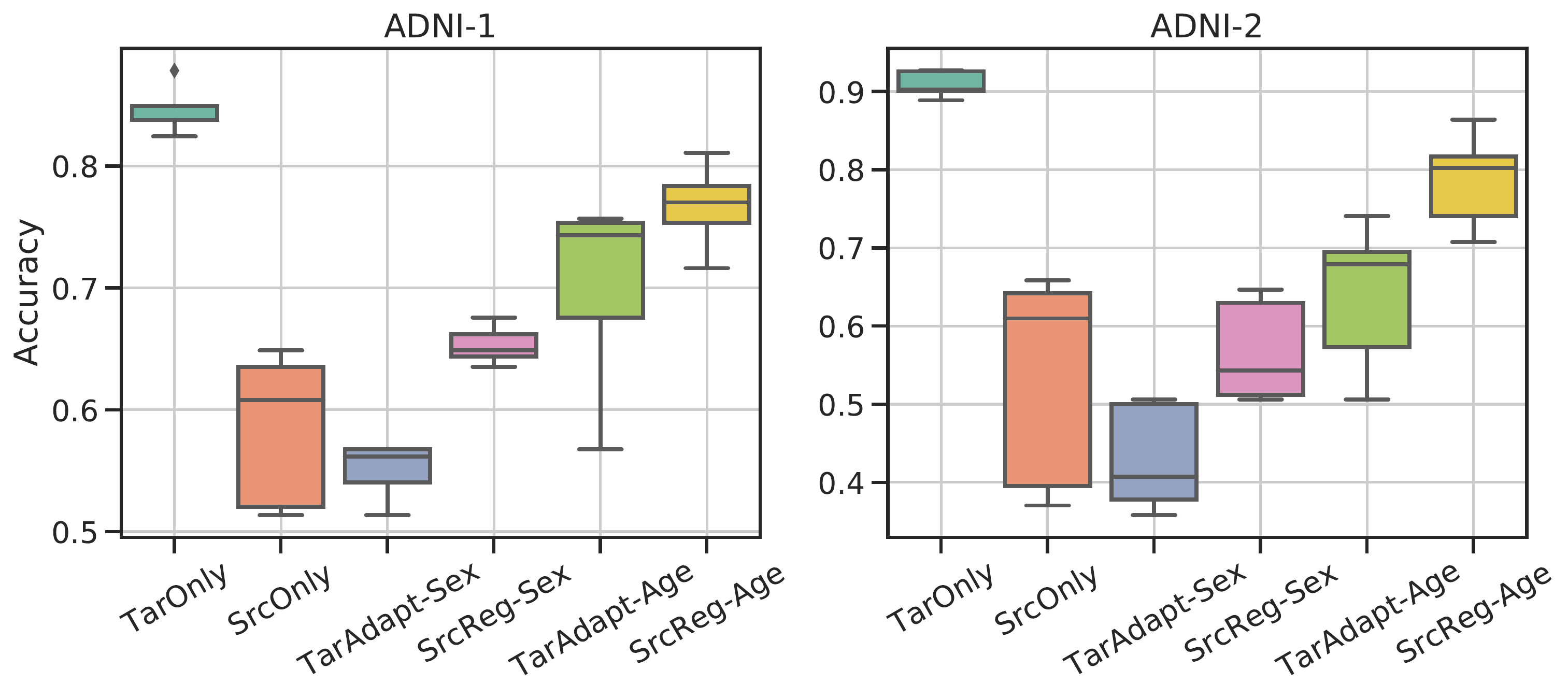}}
}
\caption{\textbf{Alzheimer’s disease classification accuracy comparison in iSTAGING consortium.} We show results with auxiliary tasks of both sex and age predictions. SrcOnly and TarOnly are vanilla models trained on source-study and target-study respectively. TarAdapt and SrcReg corresponding to the two adaptation steps in the framework. Adapted models show substantial improvements.}
\label{fig:istaging_result}
\end{figure}

\subsection{Inter-Study Alzheimer’s Disease Classification}
\label{sec:cross-ad}

\begin{table}[!t]
\caption{\textbf{Inter-study Alzheimer’s disease classification accuracy (\%) for the iSTAGING consortium data.} We report the mean accuracy and standard deviation (in round brackets) across 5-fold cross-validation$^1$.
We use sex classification and age regression as auxiliary tasks separately. TarOnly and SrcOnly denote models trained on target-study and source-study respectively. JointSup represents training on both primary and auxiliary tasks. TarAdapt and SrcReg are adapted models from the second and third phases of the proposed method. Adapted models achieve higher accuracies and reduced variance compared to SrcOnly models.}
\label{tab:istaging_result}
\begin{small}
\begin{center}
\begingroup
\setlength{\tabcolsep}{3pt}
\resizebox{\linewidth}{!}{
\begin{tabular}{l cccccccc| ccc|ccc }
\toprule
\multirow{2}{*}{ } & \multirow{2}{*}{Study}  & \multirow{2}{*}{TarOnly} & \multirow{2}{*}{SrcOnly} & \multirow{2}{*}{SVM} & \multirow{2}{*}{IRM} & \multirow{2}{*}{ADDA} & \multirow{2}{*}{DANN} & \multirow{2}{*}{Fusion} & \multicolumn{3}{c}{Sex} & \multicolumn{3}{c}{Age} \\
& & & & & & & & & JointSup & TarAdapt & SrcReg & JointSup & TarAdapt & SrcReg \\
\midrule
\multirow{2}{*}{Source} & \multirow{2}{*}{ADNI-1} & \multirow{2}{*}{-} & 84.55 & 86.99 & 86.98 & \multirow{2}{*}{-} & \multirow{2}{*}{-} & 84.51 & 84.54 & \multirow{2}{*}{-}  & \multirow{2}{*}{-}
& 78.60 & \multirow{2}{*}{-} & \multirow{2}{*}{-} \\
& & & (1.82) & (1.67) & (4.36) & & & (5.97) & (4.07) & & & (9.47) \\
\multirow{2}{*}{Target} & \multirow{2}{*}{ADNI-2} & 90.91 & 86.24
& 83.54 & 88.95 & 83.64 & 86.36 & 85.48 & 80.31 & 77.13 & 87.22
& 80.14  & 88.70  & \bf{89.68} \\
& & (1.49) & (6.65) & (4.44) & (2.96) & (2.65) & (3.27) & (3.14) & (7.53) & (8.82) & (4.03) & (8.58) & (3.13) & \bf{(1.47)} \\
\multirow{2}{*}{Target} & \multirow{2}{*}{PENN} & 92.14 & 78.84
& 81.72 & 80.59 & 76.36 & 79.21 & 80.39 & 82.69 & 83.92 & 86.89
& 78.68 & 87.93 & \bf{89.33} \\
& & (2.76) & (5.63) & (2.47) & (4.15) & (3.07) & (2.55) & (4.15) & (1.53) & (3.69) & (3.26) & (4.31) & (3.98) & \bf{(3.01)} \\
\midrule
\multirow{2}{*}{Source} & \multirow{2}{*}{ADNI-2} & \multirow{2}{*}{-} & 90.91 & 86.49 & 89.43 & \multirow{2}{*}{-} & \multirow{2}{*}{-} & 88.95 & 89.67 & \multirow{2}{*}{-}  & \multirow{2}{*}{-}
& 90.91 & \multirow{2}{*}{-} & \multirow{2}{*}{-} \\
& & & (1.49) & (4.39) & (2.53) & & & (2.18) & (5.34) & & & (3.15) \\
\multirow{2}{*}{Target} & \multirow{2}{*}{ADNI-1} & 84.55 & 82.92
& 79.68 & 82.92 & 81.74 & 82.94 & 83.76 & 80.21 & 80.21  & \bf{85.08}
& 82.38  & 83.47  & 84.82 \\
& & (1.82) & (5.14) & (2.21) & (4.25) & (1.31) & (3.75) & (4.88) & (2.66) & (3.16) & \bf{(3.03)} & (4.48) & (6.54) & (5.59) \\
\multirow{2}{*}{Target} & \multirow{2}{*}{PENN} & 92.14 & 79.54
& 68.47 & 80.81 & 75.40 & 80.37 & 76.93 & 78.30 & 78.16 & 83.22
& 80.59 & 82.86 & \bf{84.78} \\
& & (2.76) & (2.46) & (6.37) & (3.27) & (3.26) & (4.74) & (7.56) & (6.14) & (6.01) & (4.76) & (4.44) & (2.90) & \bf{(4.88)} \\
\midrule
\multirow{2}{*}{Source} & \multirow{2}{*}{PENN} & \multirow{2}{*}{-} & 92.14 & 90.85 & 91.96 & \multirow{2}{*}{-} & \multirow{2}{*}{-} & 92.48 & 91.26 & \multirow{2}{*}{-}  & \multirow{2}{*}{-}
& 92.67 & \multirow{2}{*}{-} & \multirow{2}{*}{-} \\
& & & (2.76) & (1.47) & (2.83) & & & (0.89) & (2.33) & & & (2.61) \\
\multirow{2}{*}{Target} & \multirow{2}{*}{ADNI-1} & 84.55 & 58.52
& \bf{78.86} & 58.79 & 59.27 & 66.37 & 59.07 & 53.39 & 65.14 & 74.59
& 57.73  & 65.07 & 76.16 \\
& & (1.82) & (5.72) & \bf{(1.85)} & (3.51) & (4.24) & (5.75) & (5.02) & (1.72) & (5.17) & (3.40) & (6.10) & (6.47) & (4.23) \\
\multirow{2}{*}{Target} & \multirow{2}{*}{ADNI-2} & 90.91 & 53.51
& 69.79 & 49.86 & 54.83 & 63.85 & 52.08 & 38.82 & 61.78 & 73.34
& 47.94 & 66.34 & \bf{80.85} \\
& & (1.49) & (12.57) & (2.30) & (6.05) & (3.86) & (3.85) & (10.01) & (4.22) & (4.90) & (3.54) & (9.45) & (4.28) & \bf{(2.89)} \\
\bottomrule
\end{tabular}
}
\endgroup
\end{center}
\footnotesize{$^1$ k-fold CV estimated variances could be biased as studied in~\citep{bengio2004no, bates2021cross}.}
\end{small}
\end{table}

We next demonstrate inter-study domain generalization to tackle domain shift.
We use auxiliary task that are different from the primary task for
training on the source study. In this work, we consider two supervised
auxiliary tasks: sex classification and age prediction; the rationale for picking
these auxiliary tasks is that they are both tasks where ground-truth labels are
readily accessible. The pretrained feature extractor is then adapted with these
auxiliary tasks on data from the target study. Next, the pre-trained classifier for
primary task is regularized on features from the source-study extracted from the
adapted feature extractor. Finally, we test the model with the adapted
feature extractor and primary task classifier on data from the target study.

As shown in~\cref{tab:istaging_result}, we perform AD classification on ADNI-1, ADNI-2 and PENN studies with one study as source study and the other two studies as target study respectively. For comparison, we train vanilla models (SrcOnly) on the source-studies and test the model on the target studies. We find that there are very large gaps between the testing accuracies of SrcOnly classifiers on target-studies and the validation results of the TarOnly models that are directly trained on the target-studies, highlighting poor generalization. For example, a SrcOnly model trained on ADNI-1 shows 78.84\% accuracy when tested on PENN, which is much lower than the TarOnly accuracy 92.14\%.
In contrast, \emph{our auxiliary task-assisted adaptation improves test accuracies substantially on all source-target settings for every auxiliary task.} We also report the comparison graphically in~\cref{fig:istaging_result} and observe that our proposed models show higher accuracies and reduced uncertainty in target-study prediction compared to SrcOnly models. Most noticeably, auxiliary task with age regression outperforms the one with sex classification in nearly all setups. We believe the reason for this is that age regression is more challenging task than a binary classification task like sex classification where the classifier need not be changed much.

\subsection{Inter-Study Schizophrenia Classification}

\begin{figure}[!tp]
\centering
{
\subfloat[Train on Penn and test on China and Munich.]{\includegraphics[width=0.9\textwidth]{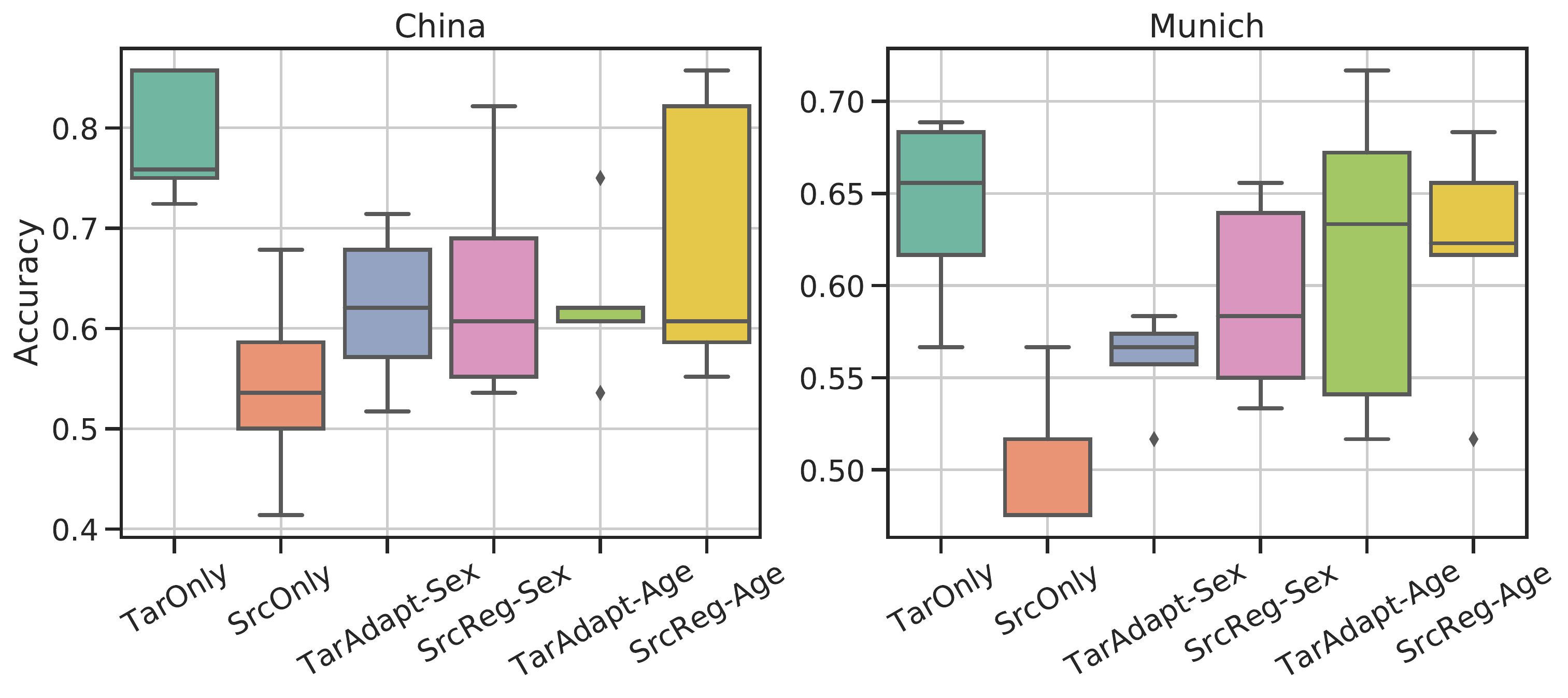}}\\
\subfloat[Train on China and test on Penn and Munich.]{\includegraphics[width=0.9\textwidth]{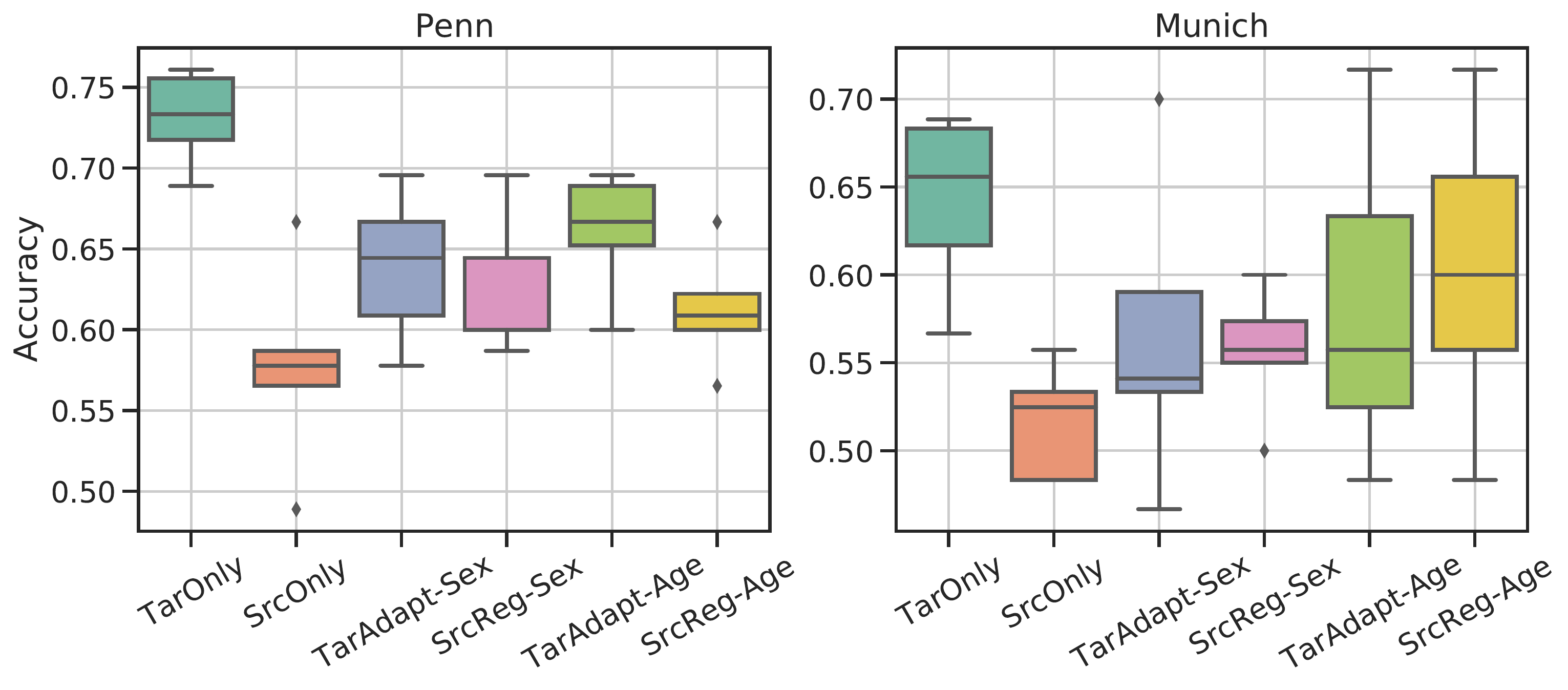}}\\
\subfloat[Train on Munich and test on Penn and China.]{\includegraphics[width=0.9\textwidth]{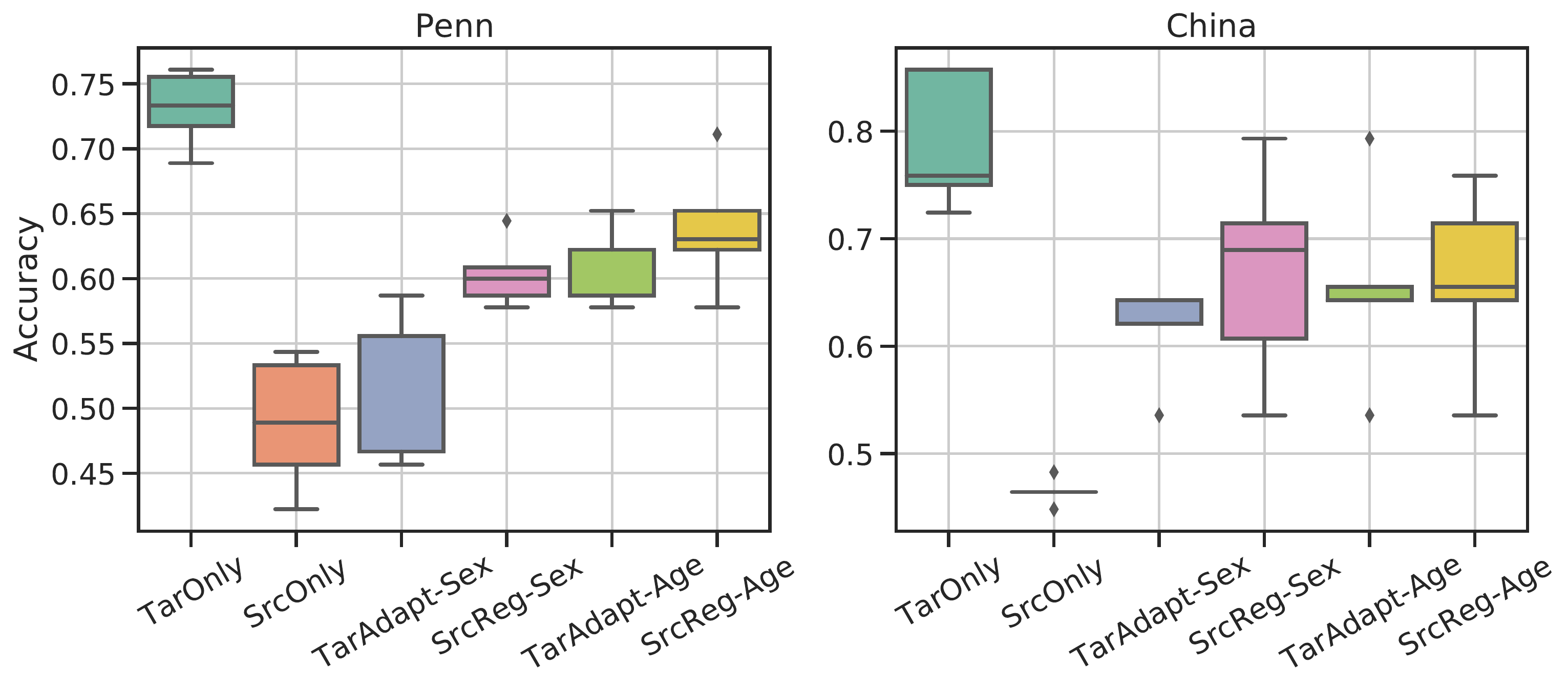}}
}
\caption{\textbf{Schizophrenia classification accuracy comparison in PHENOM consortium.} We show results with auxiliary tasks of both sex and age predictions. SrcOnly and TarOnly are vanilla models trained on source-study and target-study respectively. TarAdapt and SrcReg corresponding to the two adaptation steps in the framework. Adapted models show substantial improvements.}
\label{fig:phenom_result}
\end{figure}

\begin{table}[!t]
\caption{\textbf{Inter-study Schizophrenia classification accuracy (\%) comparison in PHENOM consortium.} We report the mean accuracy and standard deviation (in round brackets) across 5-fold cross-validation. We use sex classification and age regression as auxiliary tasks separately. TarOnly and SrcOnly denote models trained on target-study and source-study respectively. JointSup represents training on both primary and auxiliary tasks. TarAdapt and SrcReg are adapted models from the second and third phases of the proposed method. Adapted models achieve higher accuracies and reduced variance compared to SrcOnly models.}
\label{tab:phenom_result}
\begin{center}
\begin{small}
\begingroup
\setlength{\tabcolsep}{3pt}
\resizebox{\linewidth}{!}{
\begin{tabular}{l cccccccc| ccc|ccc }
\toprule
\multirow{2}{*}{ } & \multirow{2}{*}{Study}  & \multirow{2}{*}{TarOnly} & \multirow{2}{*}{SrcOnly} & \multirow{2}{*}{SVM} & \multirow{2}{*}{IRM} & \multirow{2}{*}{ADDA} & \multirow{2}{*}{DANN} & \multirow{2}{*}{Fusion} & \multicolumn{3}{c}{Sex} & \multicolumn{3}{c}{Age} \\
& & & & & & & & & JointSup & TarAdapt & SrcReg & JointSup & TarAdapt & SrcReg \\
\midrule
\multirow{2}{*}{Source} & \multirow{2}{*}{Penn} & \multirow{2}{*}{-} & 73.12 & 66.96 & 71.37 & \multirow{2}{*}{-} & \multirow{2}{*}{-} & 73.13 & 73.15 & \multirow{2}{*}{-}  & \multirow{2}{*}{-}
& 74.46 & \multirow{2}{*}{-} & \multirow{2}{*}{-} \\
& & & (2.63) & (2.00) & (1.21) & & & (3.17) & (4.54) & & & (3.46) \\
\multirow{2}{*}{Target} & \multirow{2}{*}{China} & 78.94 & 54.29
& 65.52 & 51.40 & 53.84 & 58.30 & 57.00 & 55.67 & 61.97 & 66.97
& 53.55 & 62.02 & \bf{68.37} \\
& & (5.65) & (8.81) & (8.62) & (5.57) & (2.75) & (5.74) & (4.41) & (8.41) & (4.08) & (11.15) & (2.94) & (6.94) & \bf{(5.46)} \\
\multirow{2}{*}{Target} & \multirow{2}{*}{Munich} & 64.22 & 51.02
& 62.22 & 48.67 & 50.74 & 53.74 & 51.99 & 46.68 & 54.93 & 62.58
& 55.30 & 56.98 & \bf{62.24} \\
& & (4.56) & (3.74) & (4.99) & (2.11) & (2.36) & (2.84) & (2.68) & (3.57) & (6.02) & (4.86) & (4.21) & (3.91) & \bf{(4.26)} \\
\midrule
\multirow{2}{*}{Source} & \multirow{2}{*}{China} & \multirow{2}{*}{-} & 78.94 & 75.27 & 77.51 & \multirow{2}{*}{-} & \multirow{2}{*}{-} & 76.76 & 73.99 & \multirow{2}{*}{-}  & \multirow{2}{*}{-}
& 74.01 & \multirow{2}{*}{-} & \multirow{2}{*}{-} \\
& & & (5.65) & (7.68) & (4.47) & & & (5.76) & (3.19) & & & (4.84) \\
\multirow{2}{*}{Target} & \multirow{2}{*}{Penn} & 73.12 & 57.71
& \bf{68.77} & 55.04 & 57.56 & 58.30 & 59.45 & 51.55 & 57.67 & 63.02
& 56.84 & 66.50 & 67.38 \\
& & (2.63) & (5.66) & \bf{(5.33)} & (6.51) & (3.85) & (3.26) & (1.50) & (6.46) & (5.02) & (4.24) & (5.05) & (8.08) & (4.78) \\
\multirow{2}{*}{Target} & \multirow{2}{*}{Munich} & 64.22 & 51.64
& \bf{69.18} & 54.28 & 54.38 & 59.38 & 55.63 & 49.35 & 53.93 & 55.63
& 51.34 & 59.92 & 60.63 \\
& & (4.56) & (2.91) & \bf{(5.52)} & (4.09) & (2.65) & (4.18) & (4.74) & (2.85) & (6.43) & (5.19) & (3.80) & (4.74) & (5.74) \\
\midrule
\multirow{2}{*}{Source} & \multirow{2}{*}{Munich} & \multirow{2}{*}{-} & 64.22 & 65.23 & 67.54 & \multirow{2}{*}{-} & \multirow{2}{*}{-} & 66.25 & 65.23 & \multirow{2}{*}{-}  & \multirow{2}{*}{-}
& 66.87 & \multirow{2}{*}{-} & \multirow{2}{*}{-} \\
& & & (4.56) & (3.12) & (5.47) & & & (5.26) & (3.95) & & & (6.51) \\
\multirow{2}{*}{Target} & \multirow{2}{*}{Penn} & 73.12 & 48.89
& 65.67 & 47.11 & 54.64 & 57.29 & 48.96 & 52.39 & 50.67 & 59.03
& 44.95  & 60.35 & \bf{66.16} \\
& & (2.63) & (4.57) & (5.62) & (6.66) & (4.72) & (2.74) & (5.59) & (4.88) & (3.55) & (1.65) & (3.45) & (5.69) & \bf{(3.11)} \\
\multirow{2}{*}{Target} & \multirow{2}{*}{China} & 78.94 & 46.48
& \bf{76.08} & 46.48 & 51.04 & 54.73 & 46.48 & 46.48 & 53.52 & 66.75
& 45.07 & 59.93 & 66.85 \\
& & (5.65) & (1.09) & \bf{(9.60)} & (1.09) & (3.72) & (4.83) & (0.10) & (1.09) & (5.10) & (5.19) & (2.55) & (8.83) & (3.44) \\
\bottomrule
\end{tabular}
}
\endgroup
\end{small}
\end{center}
\end{table}

Similar to the setup in~\cref{sec:cross-ad}, we also perform schizophrenia (SCZ) classification on the PHENOM consortium with three different studies: Penn, China, and Munich. Based on existing results on this data (as discussed in the following section) we believe this is a more challenging problem compared to AD classification. As shown in~\cref{tab:phenom_result} and~\cref{fig:phenom_result}, the gap in accuracy between different studies in this case are rather significant. SrcOnly classifiers achieve around $50\%$ classification accuracies when testing on target studies; this is chance accuracy for this binary classification problem. Our methods significantly increased the testing accuracies on the target-studies in all source-target settings for both auxiliary tasks compared the SrcOnly models.

\subsection{Hyper-parameter Selection}
\label{subsec:hyperparam}

\begin{figure}[!tp]
\centering
{
\subfloat[Tune auxiliary tasks (sex and age) parameter $\beta_{a}$ for feature extractor adaptation.]{\includegraphics[width=0.9\textwidth]{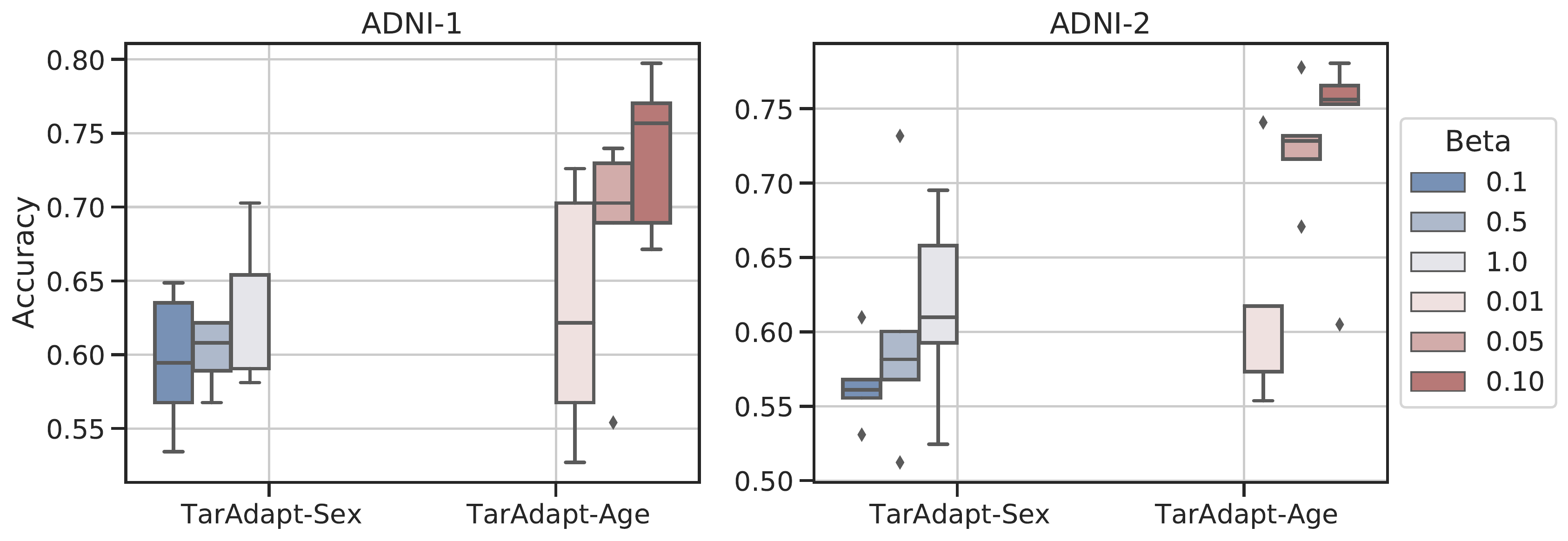}}\\
\subfloat[Tune regularization parameter $\alpha$ for primary task classifier adaptation with sex classification as the auxiliary task.]{\includegraphics[width=0.9\textwidth]{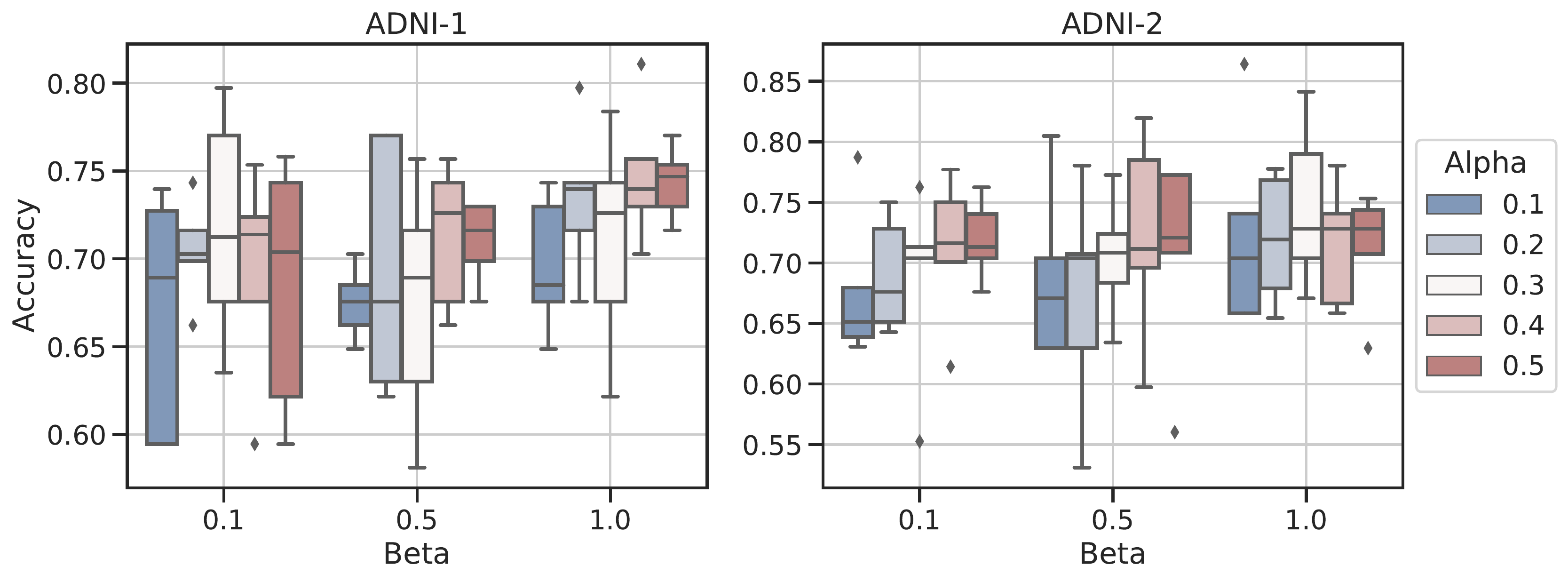}}\\
\subfloat[Tune regularization parameter $\alpha$ for primary task classifier adaptation with age regression as the auxiliary task.]{\includegraphics[width=0.9\textwidth]{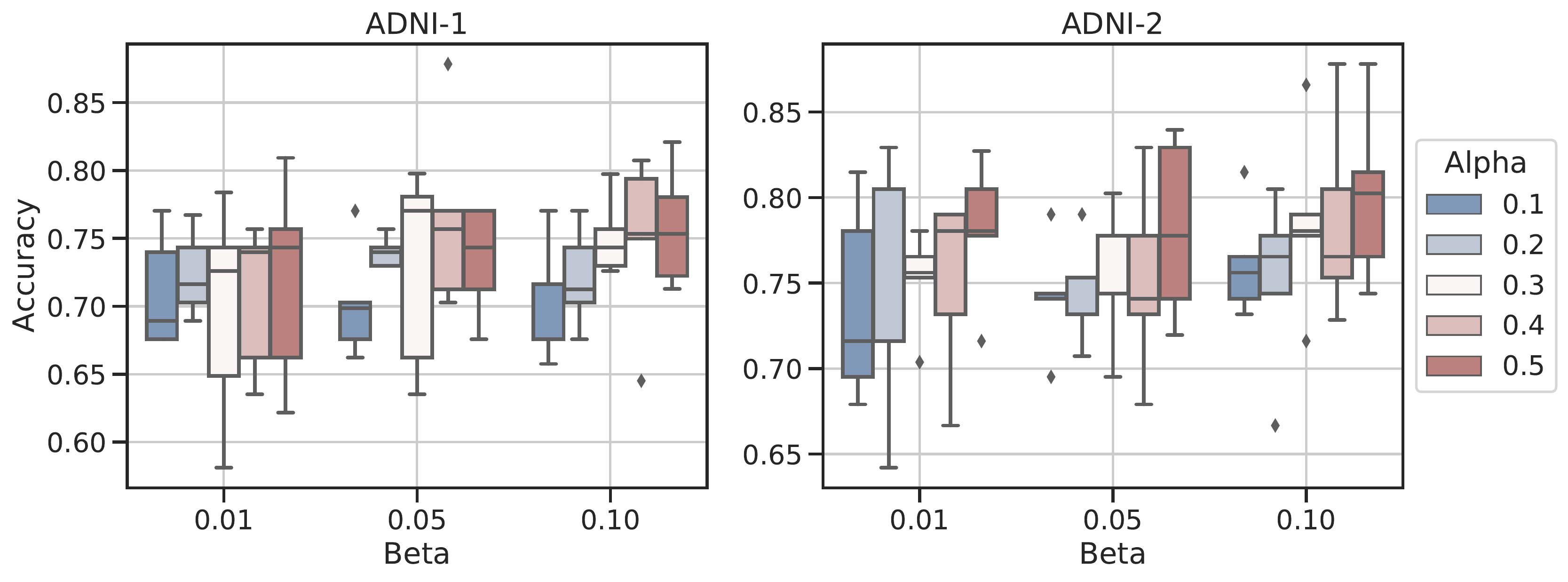}}
}
\caption{\textbf{Hyper-parameters ($\beta_{a}$ in Eqs.~\cref{eq:step_1} and $\alpha$ in Eqs.~\cref{eq:step_2}) selection case study when training on PENN and testing on ADNI-1 and ADNI-2 in iSTAGING  consortium.} In (a), we tune $\beta_a$ in range 0.1 to 1.0 for sex classification task and in range 0.01 to 0.10 for age regression task. In (b) and (c), we tune $\alpha$ in range 0.1 to 0.5 for each fixed $\beta_a$ separately in both sex and age prediction settings.}
\label{fig:param_result}
\end{figure}

We provide a case study on hyper-parameter search for AD classification on data from the iSTAGING consortium. In~\cref{fig:param_result}, we study the sensitivity of the proposed proximal term coefficient $\alpha$ in Eqs.~\cref{eq:step_2} and the auxiliary task parameter $\b_a$ in Eqs.~\cref{eq:step_1}  and we search for the optimal hyper-parameter values. We performed this study for one particular transfer experiment, namely training on PENN and adapting to data from ADNI-1 and ADNI-2 separately.
We first tune the coefficient $\b_a$ of the auxiliary tasks ([0.1, 0.5, 1.0] for sex classification and [0.01, 0.05, 0.10] for age regression). From~\cref{fig:param_result} (a), we observe that the performance of the feature extractor adaptation on the target site consistently improves as we increase the coefficient $\b_a$ for both sex and age auxiliary tasks. Age regression task is more sensitive to $\b_a$ compared to sex classification. This experiment also suggests that our chosen values of $\b_a$, i.e., 1 for sex classification and 0.1 for age regression are good.
Next, we tune coefficient $\a$ of the proposed regularization on the primary task (AD classification) classifier and search in the range [0.1, 0.5] with increments of 0.1 while fixing $\b_a$. As shown in~\cref{fig:param_result} (b) and (c), for both auxiliary tasks, for each fixed $\b_a$, the primary task adaptation of the feature extractor achieves relatively low accuracy on the target task when the coefficient $\a$ of the proximal term is small. The plots also show that the accuracy increases as $\a$ goes from 0.1 to 0.4 and accuracy eventually saturates at 0.5, which suggests that a good choice for $\a$ is 0.5.


\subsection{Traditional Machine Learning v.s. Deep Learning}

In both~\cref{tab:istaging_result} and~\cref{tab:phenom_result}, we provide a comparison of a radial basis function (RBF) kernel SVM trained on 145 brain anatomical region-of-interest (ROI) volumes extracted from the T-1 MR images with in-house tools~\citep{doshi2013multi, doshi2016muse} and 
deep learning models directly trained on target-study (TarOnly), trained on source-study and tested on target-study (SrcOnly), and our proposed model (SrcAdapt).
We observe that SVMs have essentially comparable performance, they have only slightly worse accuracy in general as compared to a deep network trained on the source study for both iSTAGING and PHENOM studies (SVM vs. SrcOnly columns). Similar observation has been reported in~\citep{koutsouleris2015individualized,rozycki2018multisite}, which suggests that traditional machine learning algorithms, such as SVM, generalize better than deep learning algorithms. However, the SVM model in general performs better on target studies as compared to SrcOnly. The differences are quite dramatic in certain cases, e.g., source PENN and target ADNI-1 in iSTAGING gets a boost of about 20\%. There are also some cases when the SVM is worse, e.g., source ADNI-2 and target PENN. As compared to this, the accuracy of SrcAdapt is consistently higher than both SrcOnly and SVM. There are some instances where this does not hold, e.g., source Munich and Target China.

%% file: journal_docs/conclusion.tex
\section{Discussion}
\label{sec:discussion}

Wide adoption of deep learning in medical imaging has been challenged by
domain shift, i.e., changes in imaging characteristics/protocols across
different sites and populations, which can significantly decrease
generalization performance of models learned in a reference dataset. In this
work, we present a systematic framework for improving the prediction accuracy
of a deep learning model on both single-study and multi-study data with shifted
distributions. In the single-study scenario, we adopt transfer learning with a
weight-constraint penalty term to fine-tune a pre-trained model with a
sub-group of data, e.g., age range, race, and scanner type. In the multi-study
scenario, we adapt a source study-trained model to the target-study with the
assistance of the auxiliary task, e.g., sex classification and age regression,
which is widely accessible in the target domain and helps capture imaging
characteristics of domain shift. By leveraging the demographic information of
each participant, we train the primary task (prediction) and auxiliary task
simultaneously with the source-study data where features are extracted for both
tasks. To generalize the model on the target-study, we fine-tune the
feature-extractor by training on the auxiliary task and regularize the
parameters associated with the primary task (prediction) by re-playing on the
source-study. From extensive experiments, we observed that our method
significantly improved the models' prediction accuracy and stability for
different age, race, and scanner groups in single-study scenarios. Additionally,
both sex and age-based auxiliary tasks helped in transferring models between
studies when labels are not available in the target-study.

\paragraph{Transfer learning and domain-shift remediation might help reduce health care disparities}

Biomedical data inequality between different age, ethnic, and scanner groups
is set to generate new health care disparities~\citep{gao2020deep}.
Insufficient training data in a particular group might lead to under-trained
deep learning model with sub-optimal prediction performance. Thus, the health
care received by the data-disadvantaged minority groups may be weakened. For
example, in ADNI dataset, there's only 19.3\% participants with age greater
than 80, whereas 51.5\% of the participants are aged between 70 and 80. The AD
classification accuracy for participants in the age group $70-80$ is $90.14\%$ and
$88.62\%$ for $>80$ age group. Biases from such data inequalities among age
groups can be partially remedied by transfer learning. The transferred model
on age group $>80$ achieved $97.10\%$ accuracy in AD classification.
Similarly, for ethnic groups, we improved the prediction accuracy in
African-American group who represented $20.8\%$ of the PENN dataset from
$95.36\%$ to $98.15\%$. On the other hand, transfer learning also helps in the
majority groups. For example, the prediction accuracy of Caucasians  ($77.4\%$
in PENN) is $90.23\%$ and a transferred model reached $97.93\%$. Similarly, in
scanner groups, Siemens Trio contributed $65.0\%$ in AIBL dataset which has an
AD classification accuracy of $90.88\%$, whereas the improved model achieved
$98.51\%$.

\paragraph{Pathology-specific classification benefits from gender and age guided features}

In the inter-study domain generalization experiments, we utilize gender and age
guided auxiliary tasks to help adapt a model trained on the source study to the
target study for two disease-specific classification tasks, namely Alzheimer’s
disease and schizophrenia classification. We observed that high-level
information like age and gender helped in extracting neuroimaging-related
features for disease classification across studies, presumably because the
domain shift in imaging features used for the auxiliary tasks was accounted
for in a way that also helped the main (disease) classification tasks. We also
find that age prediction  consistently outperforms  sex classification as an
auxiliary task, in all experimental setups for both diseases (Alzheimer’s
disease and schizophrenia). This might be due to the fact that brain aging is
closely connected to neurodegeneration. Since brain shrinkage in different
regions is associated with many diseases, age prediction is likely to
naturally help in correcting for domain shift in pathology-specific features.
This is also reported in~\citep{bashyam2020mri} where a brain age prediction
model could better capture high-level abstract neuroimaging features in
transfer learning.

\paragraph{Deep learning models generalize poorly under small sample size}

In the schizophrenia classification experiments, the PHENOM consortium
consists of 227, 142, and 302 subjects in Penn, China, and Munich datasets
separately. Since the neuropathologic patterns of schizophrenia are indistinct
and varied, and the training samples are limited, the classification task is
difficult. We observed that models trained on one study could not generalize to
another study. For example, model trained on Penn achieved $73.12\%$ prediction
accuracy but only $51.02\%$ is reported when testing on Munich. A similar
phenomenon has been reported in~\citep{bashyam2020medical} for schizophrenia
classification with neural networks on MR images. In contrast, traditional
machine learning
algorithms~\citep{koutsouleris2015individualized,rozycki2018multisite}, such
as supported vector machine (SVM), have shown adequate generalization ability
on human-designed features, e.g., regional volumetric maps
(RAVENS)~\citep{davatzikos2001voxel} that quantify gray matter volume at each
voxel, extracted from MR images. By utilizing the proposed method, we improved
the classification accuracy substantially on the target-study.

\paragraph{} The proposed framework can easily be adapted to any auxiliary tasks that are
accessible to different applications. One possible direction for future
research is to explore the generalizability of self-supervised learning,
e.g. contrastive learning, when labels are not available.

%% file: journal_docs/appendix.tex

\begin{appendix}

\section{Site differences}
\label{site}

We show the MRI scanner manufacturers, magnetic strength, and acquisition protocols, such as repetition time (TR), time to echo (TE), inversion time (TI), and field-of-view (FOV) in Table~\ref{tab:site_diff}.

\begin{table}[h!]
\caption{\tbf{Scanner manufacturers and acquisition protocols cross studies.} The scans are sampled from diverse scanner manufacturers and acquisition protocols cross-studies, which indicates the strong heterogeneous property of the data.}
\label{tab:site_diff}
\begin{center}
\begin{small}
\begingroup
\setlength{\tabcolsep}{3pt}
\begin{tabular}{l cccccc }
\toprule
Study & Scanner & Mag Strength & TR (ms) & TE (ms) & TI (ms) & FOV (mm) \\
\midrule
ADNI-1 & GE, Philips, Siemens & 1.5T & 1900 & 2.98 & 900 & $250\times 256\times 256$ \\
ADNI-2/GO & GE, Philips, Siemens & 3.0T & 2300 & 3.16 & 900 &  $208\times 240\times 256$ \\
PENN &  Siemens & 3.0T & 1600 & 3.87 & 950 & - \\
AIBL & Siemens & 1.5T, 3.0T & 1900 & 2.13 & 900 & $240\times 256\times 160$ \\
\midrule
Penn & Siemens & 3.0T & 1810 & 3.51 & 1100 & $240\times 180\times 160$ \\
China & GE & 3.0T & 8.2 & 3.2 & 450 & $256\times 256\times 188$ \\
Munich & Siemens & 1.5T & 11.6 & 4.9 & - & $230\times 256\times 126$ \\
\bottomrule
\end{tabular}
\endgroup
\end{small}
\end{center}
\end{table}

\section{Details of the neural architecture}
\label{arch}

\begin{table}[h!]
\caption{\textbf{Feature extractor network.} Padding for max-pooling layers depends on the input: columns of zeros are added along a dimension until the size along this dimension is a multiple of the stride size.}
\label{tab:arch_e}
\begin{center}
\begin{small}
\begin{tabular}{l ccccc }
\toprule
Layer & Kernel Size & Feature \# & Stride & Padding & Out Size \\
\midrule
Conv + BN + ReLU & $5\times5\times5$ & 8 & 1 & 1 & $8\times193\times229\times193$ \\
MaxPool & $2\times2\times2$ & - & 2 & adaptive & $8\times97\times115\times97$ \\
Conv + BN + ReLU & $5\times5\times5$ & 16 & 1 & 1 & $16\times97\times115\times97$ \\
MaxPool & $2\times2\times2$ & - & 2 & adaptive & $16\times49\times58\times49$ \\
Conv + BN + ReLU & $5\times5\times5$ & 32 & 1 & 1 & $32\times49\times58\times49$ \\
MaxPool & $2\times2\times2$ & - & 2 & adaptive & $32\times25\times29\times25$\\
Conv + BN + ReLU & $5\times5\times5$ & 64 & 1 & 1 & $64\times25\times29\times25$ \\
MaxPool & $2\times2\times2$ & - & 2 & adaptive & $64\times13\times15\times13$ \\
Conv + BN + ReLU & $5\times5\times5$ & 128 & 1 & 1 & $128\times13\times15\times13$ \\
MaxPool & $2\times2\times2$ & - & 2 & adaptive & $128\times7\times8\times7$ \\
\bottomrule
\end{tabular}
\end{small}
\end{center}
\end{table}

\begin{table}[h!]
\caption{\textbf{Architecture of the classifier for the primary and auxiliary task}.
The feature size (n) of the final layer is depends on the task; for sex classification the output has two logits whereas for age regression, the output is single real-valued output.}
\label{tab:arch_c}
\begin{center}
\begin{small}
\begin{tabular}{l ccc }
\toprule
Layer & Feature \# & Dropout Rate & Out Size \\
\midrule
Dropout & - & 0.5 & 19,200 \\
Linear + ReLU & 1,300 & - & 1,300 \\
Linear + ReLU & 50 & - & 50 \\
Linear & n & - & n \\
\bottomrule
\end{tabular}
\end{small}
\end{center}
\end{table}

\end{appendix}